\documentclass[10pt,twocolumn,letterpaper]{article}

\usepackage{iccv}
\usepackage{times}
\usepackage{epsfig}
\usepackage{graphicx}
\usepackage{amsmath}
\usepackage{amssymb}
\usepackage[dvipsnames]{xcolor}

\usepackage[pagebackref=true,breaklinks=true,letterpaper=true,colorlinks,bookmarks=false,citecolor=ForestGreen]{hyperref}

\usepackage[square,numbers,sort&compress]{natbib}
\usepackage[inline]{enumitem} 
\usepackage{algpseudocode}
\usepackage[font=footnotesize,labelfont=bf]{caption}
\usepackage{array}
\usepackage{multirow}
\usepackage{booktabs}
\usepackage{algorithm}
\usepackage{subcaption}
\usepackage[normalem]{ulem}
\usepackage{xparse}
\usepackage{pifont}
\usepackage{bm}
\usepackage{listings}
\usepackage{mwe} \usepackage{makecell}
\usepackage{color, colortbl}
\usepackage{cleveref}

\usepackage[scaled=0.85]{DejaVuSansMono}

\newcommand{\mytexttt}[1]{{\texttt{#1}}}

\crefformat{section}{\S#2#1#3} \crefformat{subsection}{\S~#2#1#3}
\crefformat{subsubsection}{\S~#2#1#3}
\crefformat{table}{Table~#2#1#3}
\crefformat{figure}{Figure~#2#1#3}

\newcommand{\ImNetDataset}{ImageNet\xspace}
\newcommand{\ImNet}{ImageNet-1k\xspace}
\newcommand{\ImNetFull}{ImageNet-22k\xspace}
\newcommand{\YFCCFull}{YFCC-100M\xspace}
\newcommand{\YFCC}{YFCC\xspace}
\newcommand{\YFCCone}{YFCC-1M\xspace}
\newcommand{\YFCCten}{YFCC-10M\xspace}
\newcommand{\YFCCfifty}{YFCC-50M\xspace}

\newcommand{\Places}{Places205\xspace}

\newcommand{\VOCseven}{VOC07\xspace}
\newcommand{\VOCseventwelve}{VOC07+12\xspace}

\newcommand{\COCO}{COCO2014\xspace}
\newcommand{\NYU}{NYUv2\xspace}
\newcommand{\numtasks}{9\xspace}
\newcommand{\gibson}{Gibson\xspace}

\newcommand{\alexnet}{AlexNet\xspace}

\newcommand{\resnet}{ResNet\xspace}
\newcommand{\resnetfifty}{ResNet-50\xspace}
\newcommand{\spatialbn}{\texttt{SpatialBN}\xspace}
\newcommand{\relu}{\texttt{Relu}\xspace}

\newcommand{\conv}[1]{\mytexttt{conv#1}\xspace}

\newcommand{\detectron}{\mytexttt{Detectron}\xspace}

\newcommand{\pretext}{pretext\xspace}
\newcommand{\jigsaw}{\mytexttt{Jigsaw}\xspace}

\newcommand{\permset}{\mathcal{P}}

\newcommand{\colorization}{\mytexttt{Colorization}\xspace}
\newcommand{\imageab}{\bm{Y}}
\newcommand{\imageabBinned}{\bm{Z}}

\newcommand{\imagel}{\bm{X}}
\newcommand{\colorbins}{\mathcal{Q}}
\newcommand{\numsoftbins}{K}

\newcommand{\colorizationH}{Colorization\xspace}
\newcommand{\jigsawH}{Jigsaw\xspace}

\newcommand{\image}{\bm{I}}

\newcommand{\myline}{\vspace{0.05in}}

\newlength\savewidth\newcommand\shline{\noalign{\global\savewidth\arrayrulewidth
  \global\arrayrulewidth 1pt}\hline\noalign{\global\arrayrulewidth\savewidth}}

\newlength\thinwidth\newcommand\thinline{\noalign{\global\savewidth\arrayrulewidth
  \global\arrayrulewidth 0.5pt}\hline\noalign{\global\arrayrulewidth\savewidth}}

\iccvfinalcopy

\ificcvfinal\fi
\begin{document}

\title{Scaling and Benchmarking Self-Supervised Visual Representation Learning}

\author{
Priya Goyal \quad \quad Dhruv Mahajan \quad \quad Abhinav Gupta$^{*}$ \quad \quad Ishan Misra\thanks{Equal contribution} \\
Facebook AI Research
}

\maketitle

\begin{abstract}
   Self-supervised learning aims to learn representations from the data itself without explicit manual supervision. Existing efforts ignore a crucial aspect of self-supervised learning - the ability to \textbf{scale} to large amount of data because self-supervision requires no manual labels. In this work, we revisit this principle and scale two popular self-supervised approaches to \textbf{100 million images}. We show that by scaling on various axes (including data size and problem `hardness'), one can largely match or \textbf{even exceed} the performance of supervised pre-training on a variety of tasks such as object detection, surface normal estimation (3D) and visual navigation using reinforcement learning. Scaling these methods also provides many interesting insights into the limitations of current self-supervised techniques and evaluations. We conclude that current self-supervised methods are not `hard' enough to take full advantage of large scale data and do not seem to learn effective high level semantic representations. We also introduce an \textbf{extensive benchmark} across \textbf{\numtasks different datasets and tasks}. We believe that such a benchmark along with \textbf{comparable} evaluation settings is necessary to make meaningful progress. Code is at: {{\url{https://github.com/facebookresearch/fair_self_supervision_benchmark}}}.
\end{abstract}

\section{Introduction}
\vspace{-0.05in}

Computer vision has been revolutionized by high capacity Convolutional Neural Networks (ConvNets)~\cite{lecun1989backpropagation} and large-scale labeled data (e.g., ImageNet~\cite{deng2009imagenet}). Recently~\cite{mahajan2018exploring,sun2017revisiting}, weakly-supervised training on hundreds of millions of images and thousands of labels has achieved state-of-the-art results on various benchmarks. Interestingly, even at that scale, performance increases only log-linearly with the amount of labeled data. Thus, sadly, what has worked for computer vision in the last five years has now become a bottleneck: the size, quality, and availability of supervised data.

One alternative to overcome this bottleneck is to use the self-supervised learning paradigm. In discriminative self-supervised learning, which is the main focus of this work, a model is trained on an auxiliary or `pretext' task for which ground-truth is available for free. In most cases, the pretext task involves predicting some hidden portion of the data (for example, predicting color for gray-scale images~\cite{larsson2016learning,zhang2016colorful,deshpande2015learning}). Every year, with the introduction of new pretext tasks, the performance of self-supervised methods keeps coming closer to that of ImageNet supervised pre-training. The hope around self-supervised learning outperforming supervised learning has been so strong that a researcher has even bet gelato~\cite{gelato}.

Yet, even after multiple years, this hope remains unfulfilled. Why is that? In attempting to come up with clever pretext tasks, we have forgotten a crucial tenet of self-supervised learning: {\bf scalability}. Since no manual labels are required, one can easily scale training from a million to billions of images. However, it is still unclear what happens when we scale up self-supervised learning beyond the ImageNet scale to 100M images or more. Do we still see performance improvements? Do we learn something insightful about self-supervision? Do we surpass the ImageNet supervised performance?

In this paper, we explore scalability which is a core tenet of self-supervised learning. Concretely, we scale two popular self-supervised approaches  (\jigsaw~\cite{noroozi2016unsupervised} and \colorization~\cite{zhang2016colorful}) along three axes:
\begin{enumerate}[leftmargin=*,itemsep=0em,nolistsep]
    \item \textbf{Scaling pre-training data:} We first scale up both methods to $100 \times$ more data (\YFCCFull~\cite{thomee2015yfcc100m}). We observe that low capacity models like \alexnet~\cite{krizhevsky2012imagenet} do not show much improvement with more data. This motivates our second axis of scaling.
    \item \textbf{Scaling model capacity:} We scale up to a higher capacity model, specifically ResNet-50~\cite{he2016deep}, that shows much larger improvements as the data size increases. While recent approaches \cite{doersch2017multi,kolesnikov2019revisiting,wu2018unsupervised} used models like ResNet-50 or 101, we explore the relationship between model capacity and data size which we believe is crucial for future efforts in self-supervised learning.
    \item \textbf{Scaling problem complexity:} Finally, we observe that to take full advantage of large scale data and higher capacity models, we need `harder' pretext tasks. Specifically, we scale the `hardness' (problem complexity) and observe that higher capacity models show a larger improvement on `harder' tasks.
    \end{enumerate}

Another interesting question that arises is: how does one quantify the visual representation's quality?
We observe that due to the lack of a \textbf{standardized evaluation methodology} in self-supervised learning, it has become difficult to compare different approaches and measure the advancements in the area. To address this, we propose an \textbf{extensive benchmark suite} to evaluate representations using a consistent methodology. Our benchmark is based on the following principle: a good representation (1) transfers to \emph{many} different tasks, and, (2) transfers with \emph{limited} supervision and \emph{limited} fine-tuning. We carefully choose \numtasks different tasks (Table~\ref{tab:transfer_datasets}) ranging from semantic classification/detection to 3D and actions (specifically, navigation).

Our results show that by scaling along the three axes, self-supervised learning can \textbf{outperform ImageNet supervised} pre-training using the \emph{same} evaluation setup on non-semantic tasks of Surface Normal Estimation and Navigation.
For semantic classification tasks, although scaling helps outperform previous results, the gap with supervised pre-training remains significant when evaluating fixed feature representations (without full fine-tuning). Surprisingly, self-supervised approaches are quite competitive on object detection tasks with or without full fine-tuning. For example, on the \VOCseven \textbf{detection} task, \textbf{without any bells and whistles, our performance matches} the supervised ImageNet pre-trained model.

\begin{table}[!t]
    \centering
    \scalebox{0.65}{
    \setlength\tabcolsep{1.2pt}
    \begin{tabular}{c|cc}
        \textbf{Task} & \textbf{Datasets} & \textbf{Description} \\
        \shline
        \textbf{Image classification} & & \\
                \cref{sec:image_cls_all}& \Places & \makecell{Scene classification. 205 classes.} \\
        (Linear Classifier) & \VOCseven & \makecell{Object classification. 20 classes.} \\
        & \COCO & \makecell{Object classification. 80 classes.} \\
        \thinline
                                        \makecell{\textbf{Low-shot image classification}} & & \\
        \cref{sec:lowshot}& \VOCseven & \makecell{$\leq 96$ samples per class} \\
        (Linear Classifier) & \Places & \makecell{$\leq 128$ samples per class} \\
        \thinline
        \textbf{Visual navigation} & & \\
        \cref{sec:navigation} (Fixed ConvNet)& \gibson & Reinforcement Learning for navigation. \\
        \thinline
        \textbf{Object detection} & & \\
        \cref{sec:object_detection}& \VOCseven & \makecell{20 classes.} \\
        (Frozen conv body)& \VOCseventwelve & \makecell{20 classes.} \\
        \thinline
        \textbf{Scene geometry (3D)} & & \\
        \cref{sec:surface_normals} (Frozen conv body)& \NYU & Surface Normal Estimation. \\
        \shline
    \end{tabular}}
    \vspace{-0.08in}
    \caption{\textbf{\numtasks transfer datasets and tasks} used for \textbf{Benchmarking} in \cref{sec:benchmarking}.}
    \label{tab:transfer_datasets}
    
\end{table}

\section{Related Work}

Visual representation learning without supervision is an old and active area of research. It has two common modeling approaches: generative and discriminative. A generative approach tries to model the data distribution directly. This can be modeled as maximizing the probability of reconstructing the input~\cite{vincent2008extracting,olshausen1996emergence,masci2011stacked} and optionally estimating latent variables~\cite{salakhutdinov2009deep,huang2007unsupervised} or using adversarial training~\cite{donahue2016adversarial,mescheder2017adversarial}. Our work focuses on discriminative learning.

\par
One form of discriminative learning combines clustering with hand-crafted features to learn visual representations such as image-patches~\cite{singh2012unsupervised,doersch2012makes}, object discovery~\cite{russell2006using,sivic2005discovering}.
We focus on discriminative approaches that learn representations directly from the the visual input.
A large portion of such approaches are grouped under the term `self-supervised' learning~\cite{de1994learning} in which the key principle is to automatically generate `labels' from the data. The label generation can either be domain agnostic~\cite{caron2018deep,oord2018representation,wu2018unsupervised,bojanowski2017unsupervised} or exploit structural properties of the domain, \eg, spatial structure of images~\cite{doersch2015unsupervised}. We explore the `\pretext' tasks~\cite{doersch2015unsupervised} that exploit structural information of the visual data to learn representations. These approaches can broadly be divided into two types - methods that use multi-modal information, \eg sound~\cite{owens2016ambient} and methods that use only the visual data (images, videos). Multi-modal information such as depth from a sensor~\cite{eigen2015predicting}, sound in a video~\cite{arandjelovic2017look,arandjelovic2018objects,gao2018learning,owens2016ambient}, sensors on an autonomous vehicle~\cite{agrawal2015learning,jayaraman2015learning,zhou2017unsupervised} \etc can be used to automatically learn visual representations without  human supervision. One can also use the temporal structure in a video for self-supervised methods~\cite{mobahi2009deep,hadsell2006dimensionality,fernando2017self,misra2016shuffle,luc2017predicting}. Videos can provide information about how objects move~\cite{pathak2017learning}, the relation between viewpoints~\cite{wang2017transitive,wang2015unsupervised} \etc.

In this work, we choose to scale image-based self-supervised methods because of their ease of implementation. Many \pretext tasks have been designed for images that exploit their spatial structure~\cite{doersch2015unsupervised,noroozi2016unsupervised,noroozi2017representation,noroozi2018boosting}, color information~\cite{zhang2016colorful,larsson2016learning,larsson2017colorization,deshpande2015learning}, illumination~\cite{dosovitskiy2016discriminative}, rotation~\cite{gidaris2018unsupervised} \etc. These \pretext tasks model different properties of images and have been shown to contain complementary information~\cite{doersch2017multi}.
Given the abundance of such approaches to use, in our work, we focus on two popular approaches that are simple to implement, intuitive, and diverse: \jigsaw from~\cite{noroozi2016unsupervised} and \colorization from~\cite{zhang2016colorful}. A concurrent work~\cite{kolesnikov2019revisiting} also explores multiple self-supervised tasks but their focus is on the architectural details which is complementary to ours.
\section{Preliminaries}

We briefly describe the two image based self-supervised approaches~\cite{noroozi2017representation,zhang2016colorful} that we study in this work and refer the reader to the original papers for detailed explanations. Both these methods do \emph{not} use any supervised labels. 

\subsection{\jigsaw Self-supervision}
\label{sec:jigsaw_approach}
This approach by Noroozi \etal~\cite{noroozi2016unsupervised} learns an image representation by solving jigsaw puzzles created from an input image. The method divides an input image $\image$ into $N=9$ non-overlapping square patches. A `puzzle' is then created by shuffling these patches randomly and a ConvNet is trained to predict the permutation used to create the puzzle. Concretely, each patch is fed to a $N$-way Siamese ConvNet with shared parameters to obtain patch representations. The patch representations are concatenated and used to predict the permutation used to create the puzzle. In practice, as the total number of permutations $N!$ can be large, a fixed subset $\permset$ of the total $N!$ permutations is used. The prediction problem is reduced to classification into one of $|\permset|$ classes.

\subsection{\colorization Self-supervision}
\label{sec:colorization_approach}
Zhang \etal~\cite{zhang2016colorful} learn an image representation by predicting color values of an input `grayscale' image. The method uses the CIE \emph{Lab} color space representation of an input image $\image$ and trains a model to predict the \emph{ab} colors (denoted by $\imageab$) from the input lightness \emph{L} (denoted by $\imagel$). The output \emph{ab} space is quantized into a set of discrete bins $\colorbins=313$ which reduces the problem to a $|\colorbins|$-way classification problem. The target \emph{ab} image $\imageab$ is soft-encoded into $|\colorbins|$ bins by looking at the $\numsoftbins$-nearest neighbor bins (default value $\numsoftbins$=$10$). We denote this soft-encoded target explicitly by $\imageabBinned^\numsoftbins$. Thus, each $|\colorbins|$-way classification problem has $\numsoftbins$ non-zero values. The ConvNet is trained to predict $\imageabBinned^\numsoftbins$ from the input lightness image $\imagel$.

\begin{table}[]
    \centering
    \setlength\tabcolsep{-2pt}
    \begin{tabular}{cc}
    \includegraphics[width=0.248\textwidth]{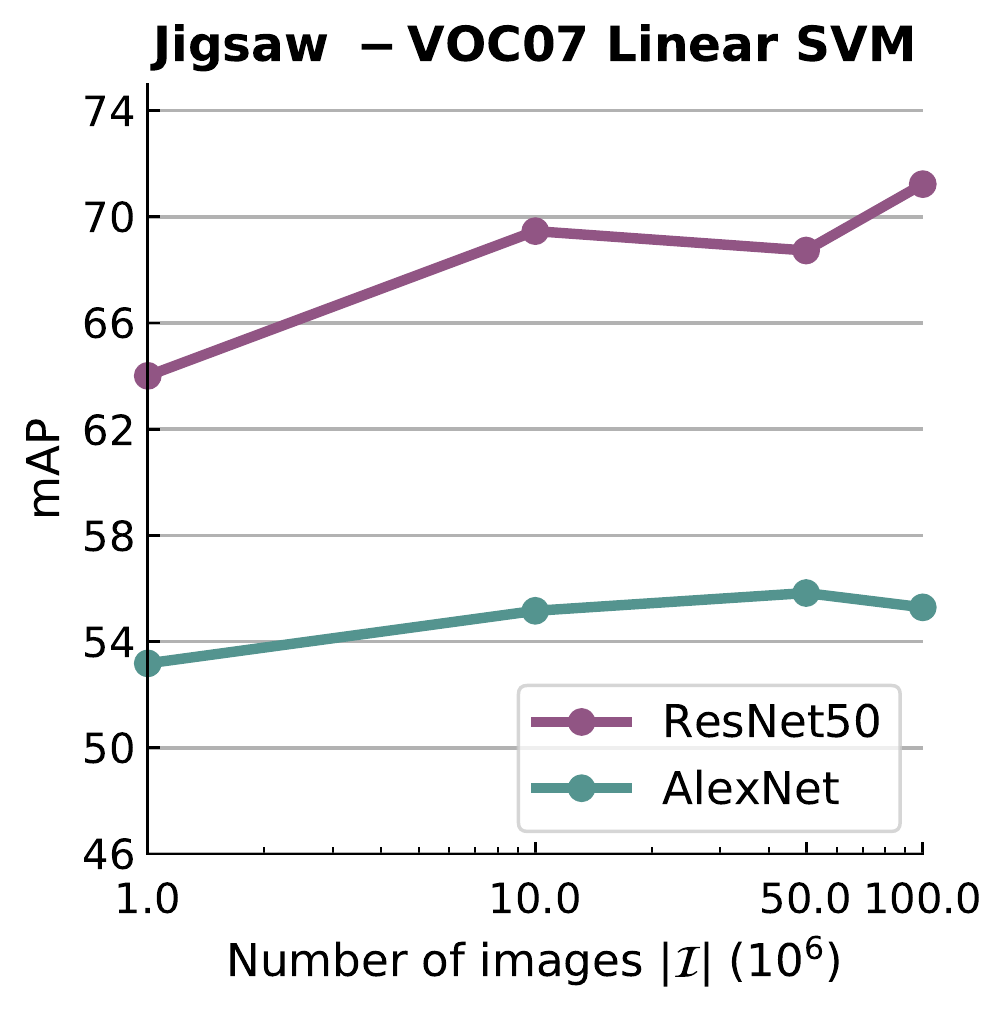} & \includegraphics[width=0.248\textwidth]{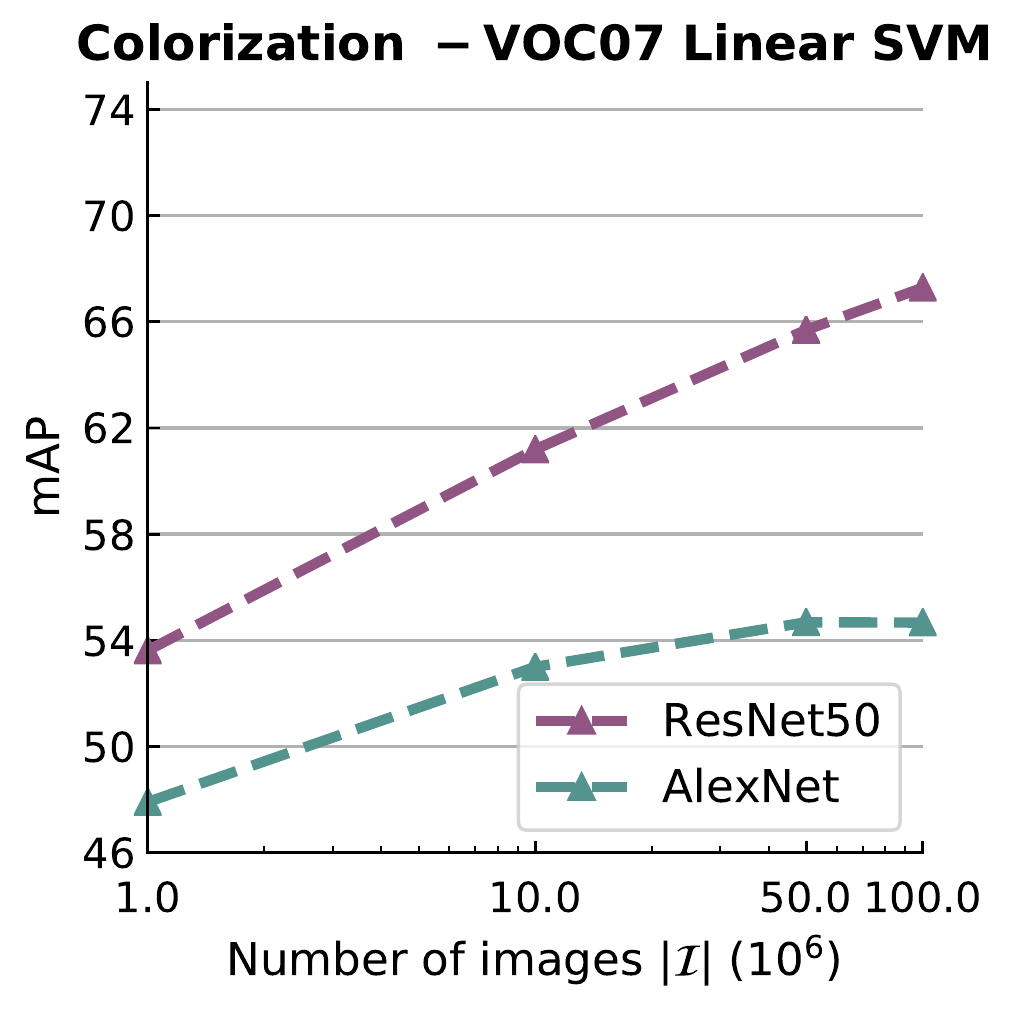}
    \end{tabular}
    \vspace{-0.2in}
    \captionof{figure}{\textbf{Scaling the Pre-training Data Size:} The transfer learning performance of self-supervised methods on the \VOCseven dataset for \alexnet and \resnetfifty as we vary the pre-training data size. We keep the problem complexity and data domain (different sized subsets of \YFCCFull) fixed. More details in \cref{sec:scaling_data}.}
    \label{fig:scaling_data}
\end{table}
\section{Scaling Self-supervised Learning}
\label{sec:scaling_ablation}
In this section, we scale up current self-supervised approaches and present the insights gained from doing so. We \emph{first} scale up the data size to $\mathbf{100\times}$ the size commonly used in existing self-supervised methods. However, observations from recent works~\cite{sun2017revisiting,mahajan2018exploring,joulin2016learning} show that higher capacity models are required to take full advantage of large datasets. Therefore, we explore the \emph{second axis} of scaling: model capacity. Additionally, self-supervised learning provides an interesting \emph{third axis}: the complexity (hardness) of \pretext tasks which can control the quality of the learned representations. 

Finally, we observe the relationships between these three axes: whether the performance improvements on each of the axes are complementary or they encompass one other. To study this behavior, we introduce a simple investigation setup. Note that this setup is different from the extensive evaluation benchmark we propose in \cref{sec:benchmarking}.

\myline
\par \noindent \textbf{Investigation Setup:} We use the task of image classification on PASCAL VOC2007~\cite{everingham2010pascal} (denoted as \VOCseven). We train linear SVMs~\cite{boser1992training} (with $3$-fold cross validation to select the cost parameter) on fixed feature representations obtained from the ConvNet (setup from~\cite{owens2016ambient}). Specifically, we choose the best performing layer: \conv{4} layer for \alexnet and the output of the last \texttt{res4} block (notation from~\cite{girshick2018detectron}) for ResNet-50. We train on the \mytexttt{trainval} split and report mean Average Precision (mAP) on the \mytexttt{test} split.

\begin{table}[!t]
    \centering
    \footnotesize{
    \setlength\tabcolsep{1.5pt}
    \begin{tabular}{l|c}
        \textbf{Symbol} & \textbf{Description} \\
        \shline
        YFCC-$X$M & \makecell{Images from the \YFCCFull~\cite{thomee2015yfcc100m} dataset. \\
        We use subsets of size $X \in [1M, 10M, 50M, 100M]$.} \\
        \thinline
                        \ImNetFull & \makecell{The full \ImNetDataset dataset (22k classes, $14M$ images)~\cite{deng2009imagenet}.}\\
        \thinline
        \ImNet & \makecell{ILSVRC2012 dataset ($1k$ classes, $1.28M$ images)~\cite{ILSVRC15}.} \\
        \thinline
    \end{tabular}
    \vspace{-0.1in}
    \caption{A list of \textbf{self-supervised pre-training datasets} used in this work. We train \alexnet~\cite{krizhevsky2012imagenet} and \resnetfifty~\cite{he2016deep} on these datasets.}
    \label{tab:pretrain_datasets}
    }
\end{table}

\subsection{Axis 1: Scaling the Pre-training Data Size}
\label{sec:scaling_data}
The first premise in self-supervised learning is that it requires `no labels' and thus can make use of large datasets. But do the current self-supervised approaches benefit from increasing the pre-training data size? We study this for both the \jigsaw and \colorization methods. Specifically, we train on various subsets (see Table~\ref{tab:pretrain_datasets}) of the \YFCCFull dataset - \YFCC-$[1, 10, 50, 100]$ million images. These subsets were collected by randomly sampling respective number of images from the \YFCCFull dataset. We specifically create these \YFCC subsets so we can keep the data domain fixed. Further, during the self-supervised pre-training, we keep other factors that may influence the transfer learning performance such as the model, the problem complexity ($|\permset|=2000$, $K=10$) \etc fixed. This way we can isolate the effect of data size on performance. We provide training details in the supplementary material.

\myline
\par \noindent \textbf{Observations:} We report the transfer learning performance on the \VOCseven classification task in \cref{fig:scaling_data}. We see that increasing the size of pre-training data improves the transfer learning performance for both the \jigsaw and \colorization methods on \resnetfifty and \alexnet. We also note that the \jigsaw approach performs better compared to \colorization. Finally, we make an interesting observation that the performance of the \jigsaw model saturates (log-linearly) as we increase the data scale from $1$M to $100$M.

\subsection{Axis 2: Scaling the Model Capacity}
\label{sec:scaling_model_capacity}

We explore the relationship between model capacity and self-supervised representation learning. Specifically, we observe this relationship in the context of the pre-training dataset size. For this, we use \alexnet and the higher capacity \resnetfifty~\cite{he2016deep} model to train on the same pre-training subsets from \cref{sec:scaling_data}.

\myline
\par \noindent \textbf{Observations:} \cref{fig:scaling_data} shows the transfer learning performance on the \VOCseven classification task for \jigsaw and \colorization approaches. We make an important observation that the performance gap between \alexnet and \resnetfifty (as a function of the pre-training dataset size) keeps \emph{increasing}. This suggests that higher capacity models are needed to take full advantage of the larger pre-training datasets.

\begin{table}[]
    \centering
    \setlength\tabcolsep{-2pt}
    \begin{tabular}{cc}
    \includegraphics[width=0.248\textwidth]{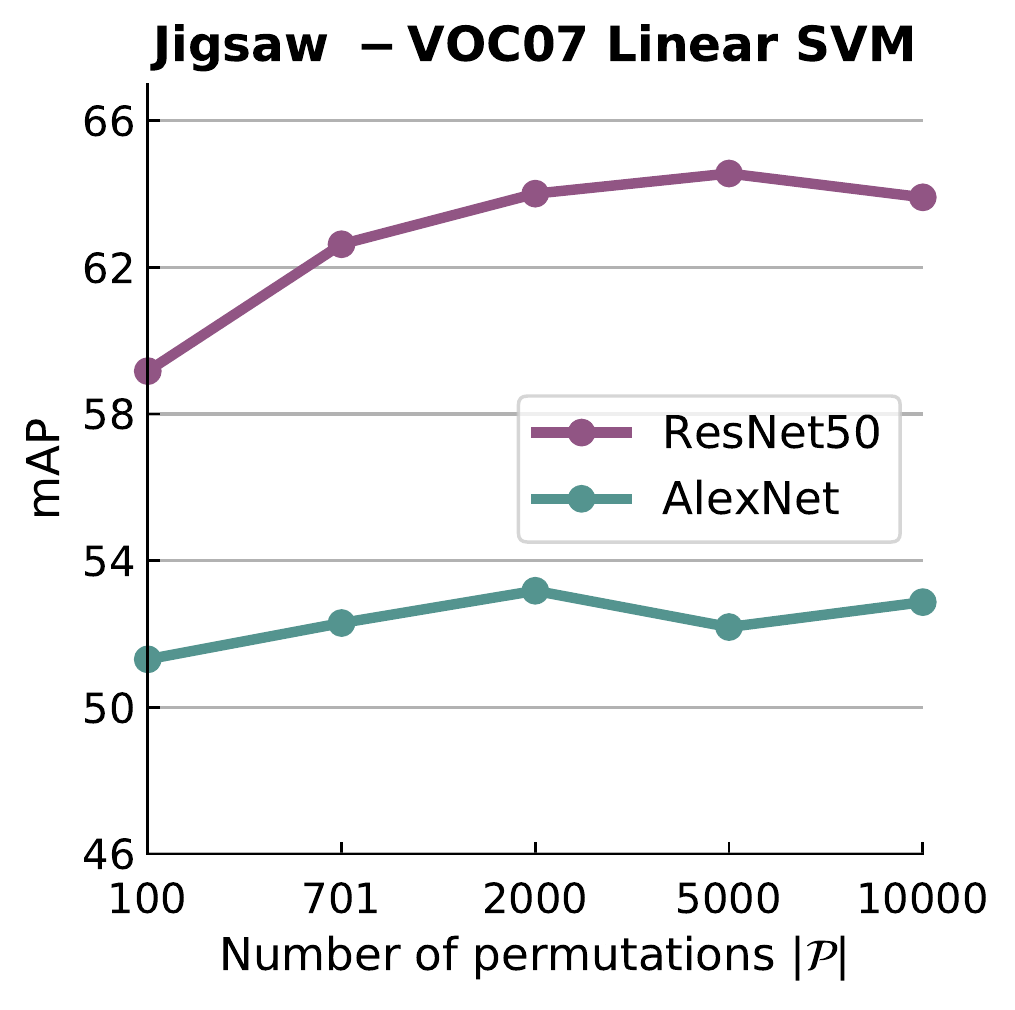} & \includegraphics[width=0.248\textwidth]{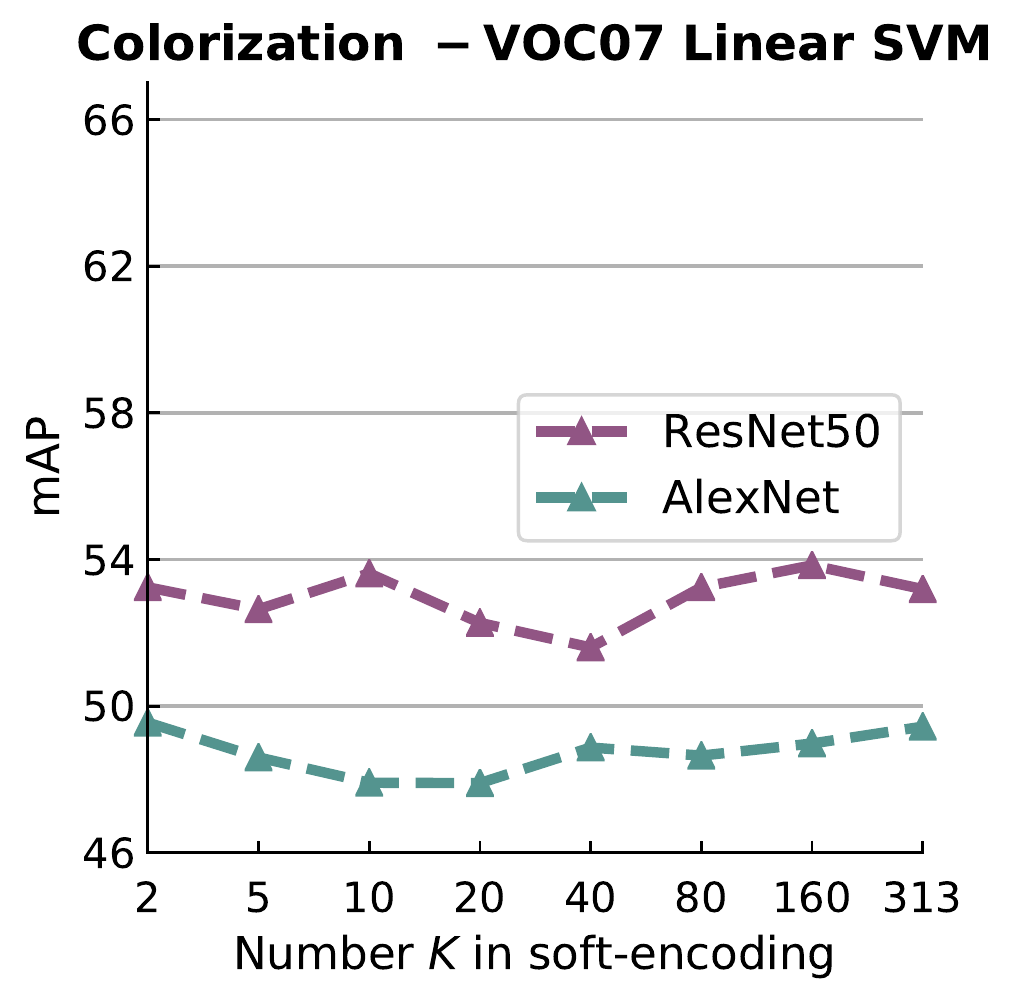} 
    \end{tabular}
    \vspace{-0.2in}
    \captionof{figure}{\textbf{Scaling Problem Complexity:} We evaluate transfer learning performance of \jigsaw and \colorization approaches on \VOCseven dataset for both \alexnet and \resnetfifty as we vary the problem complexity. The pre-training data is fixed at \YFCCone (\cref{sec:scaling_problem}) to isolate the effect of problem complexity.}
    \label{fig:scaling_problem}
\end{table}

\subsection{Axis 3: Scaling the Problem Complexity}
\label{sec:scaling_problem}
We now scale the problem complexity (`hardness') of the self-supervised approaches. We note that it is important to understand how the complexity of the \pretext tasks affects the transfer learning performance.

\myline
\par \noindent \textbf{Jigsaw:}
The number of permutations $|\permset|$ (\cref{sec:jigsaw_approach}) determines the number of puzzles seen for an image. We vary the number of permutations $|\permset| \in [100,\ 701,\ 2k,\ 5k,\ 10k]$ to control the problem complexity. Note that this is a $\bm{10\times}$ increase in complexity compared to~\cite{noroozi2016unsupervised}.\myline
\par \noindent \textbf{Colorization:} We vary the number of nearest neighbors $\numsoftbins$ for the soft-encoding (\cref{sec:colorization_approach}) which controls the hardness of the colorization problem. 
\myline
\par \noindent To isolate the effect of problem complexity, we fix the pre-training data at \YFCCone. We explore additional ways of increasing the problem complexity in the supplementary material.

\myline
\par \noindent \textbf{Observations:} We report the results on the \VOCseven classification task in \cref{fig:scaling_problem}.  For the \jigsaw approach, we see an improvement in transfer learning performance as the size of the permutation set increases. \resnetfifty shows a \textbf{5 point} mAP improvement while \alexnet shows a smaller 1.9 point improvement. The \colorization approach appears to be less sensitive to changes in problem complexity. We see $\sim$2 point mAP variation across different values of $K$. We believe one possible explanation for this is in the structure encoded in the representation by the \pretext task. For \colorization, it is important to represent the relationship between the semantic categories and their colors, but fine-grained color distinctions do not matter as much. On the other hand, \jigsaw encodes more spatial structure as the problem complexity increases which may matter more for downstream transfer task performance.

\begin{table}[!t]
    \centering
    \setlength\tabcolsep{-3pt}
    \begin{tabular}{@{}cc@{}}
    \includegraphics[width=0.248\textwidth]{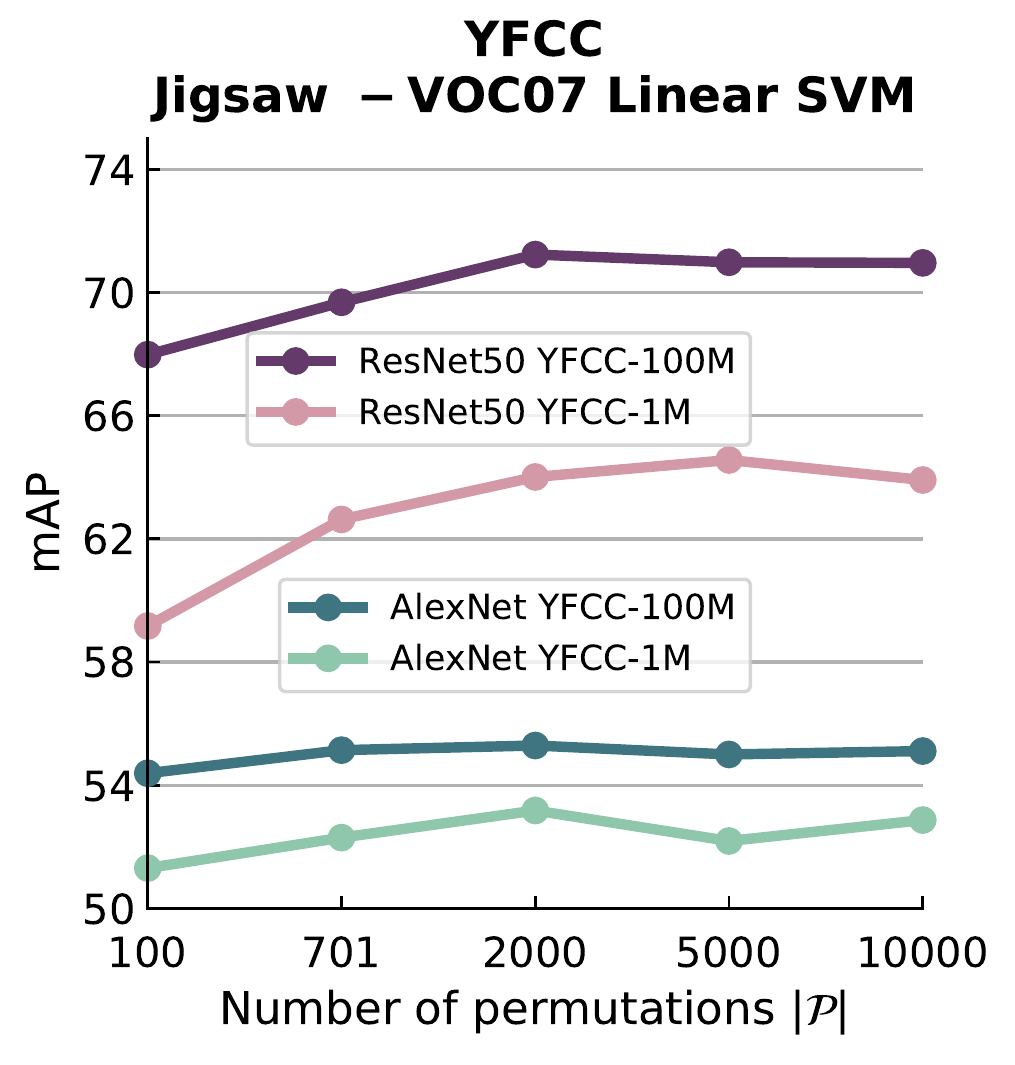} & \includegraphics[width=0.248\textwidth]{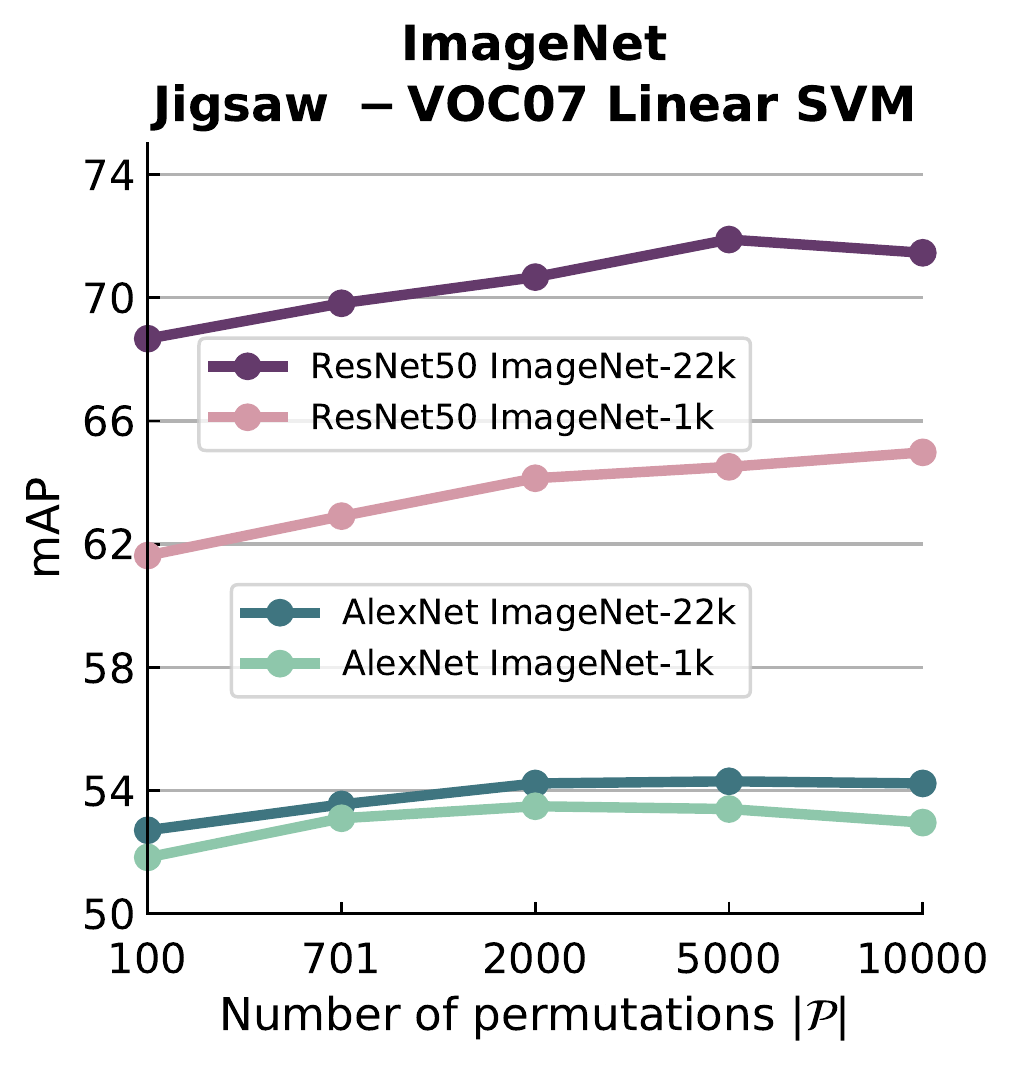} 
    \end{tabular}
    \vspace{-0.2in}
    \captionof{figure}{\textbf{Scaling Data and Problem Complexity:} We vary the pre-training data size and \jigsaw problem complexity for both \alexnet and \resnetfifty models.  We pre-train on two datasets: \ImNetDataset and \YFCC and evaluate transfer learning performance on \VOCseven dataset.}
    \label{fig:scaling_problem_and_data}
\end{table}

\subsection{Putting it together}
Finally, we explore the relationship between all the three axes of scaling. We study if these axes are orthogonal and if the performance improvements on each axis are complementary. We show this for \jigsaw approach only as it outperforms the \colorization approach consistently. Further, besides using \YFCC subsets for \pretext task training (from \cref{sec:scaling_data}), we also report self-supervised results for \ImNetDataset datasets (without using any labels). \cref{fig:scaling_problem_and_data} shows the transfer learning performance on \VOCseven task as function of data size, model capacity and problem complexity. 

We note that transfer learning performance increases on all three axes, \ie, increasing problem complexity still gives performance boost on \resnetfifty even at $100$M data size. Thus, we conclude that the three axes of scaling are complementary. We also make a crucial observation that the performance gains for increasing problem complexity are almost negligible for \alexnet but significantly higher for \resnetfifty. This indicates that we need higher capacity models to exploit hardness of self-supervised approaches.

{
\begin{table*}
    \centering
    \footnotesize{
    \setlength\tabcolsep{1.5pt}
    \renewcommand{\arraystretch}{0.6}    \begin{tabular}{cccc}
    \includegraphics[width=0.24\textwidth]{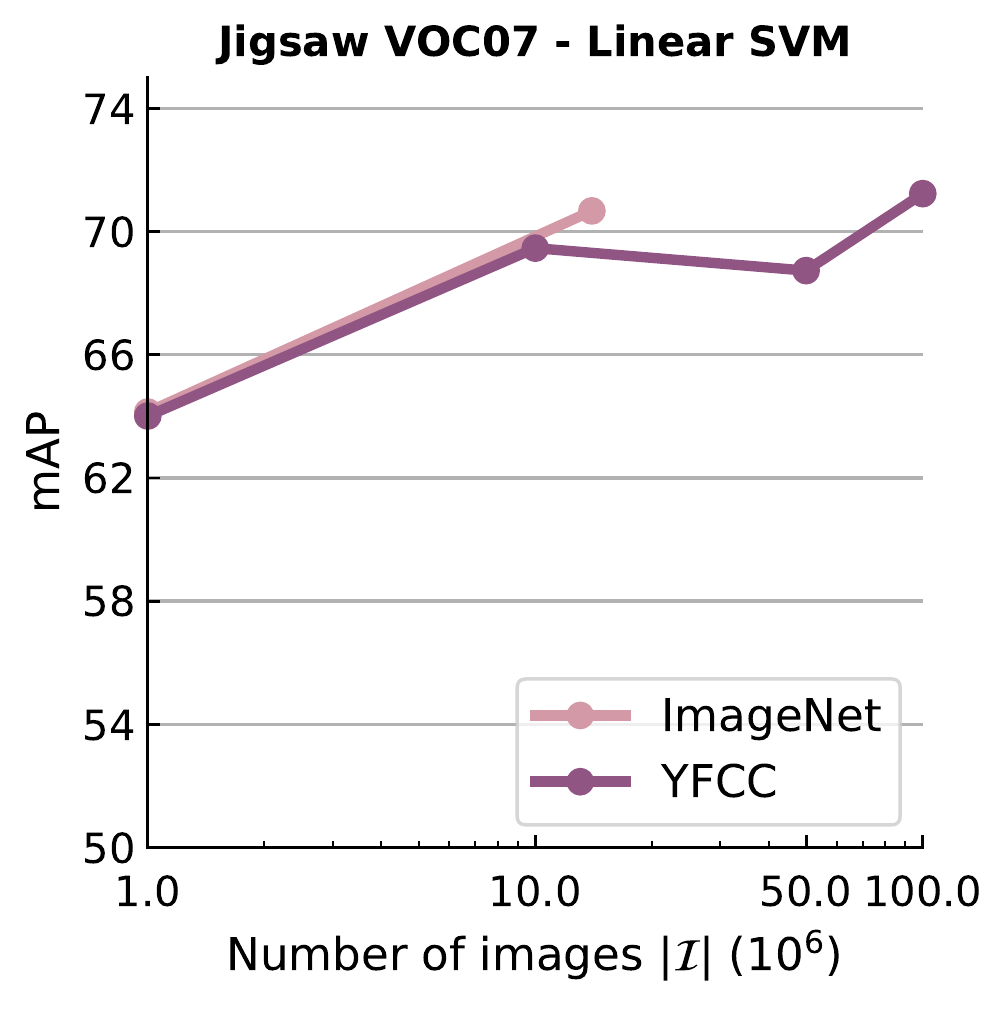} &
    \includegraphics[width=0.24\textwidth]{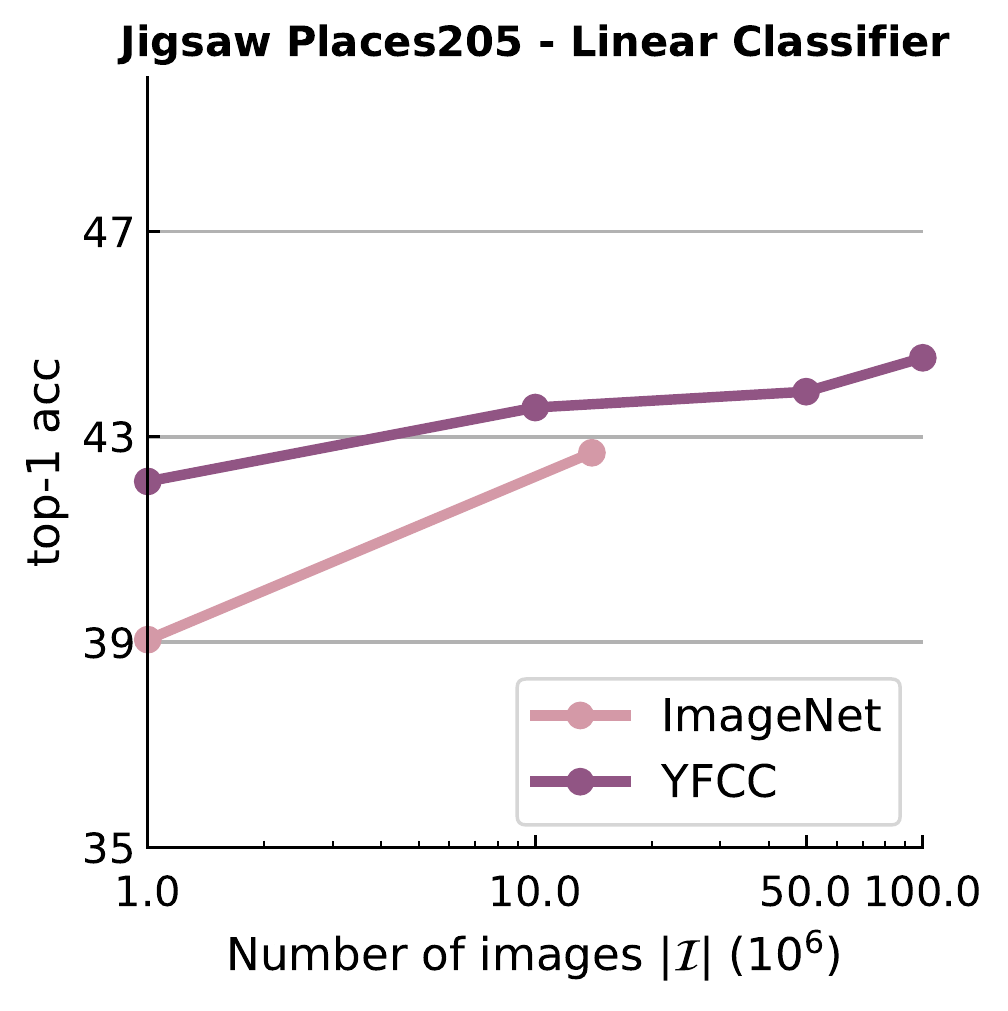}  &
    \includegraphics[width=0.24\textwidth]{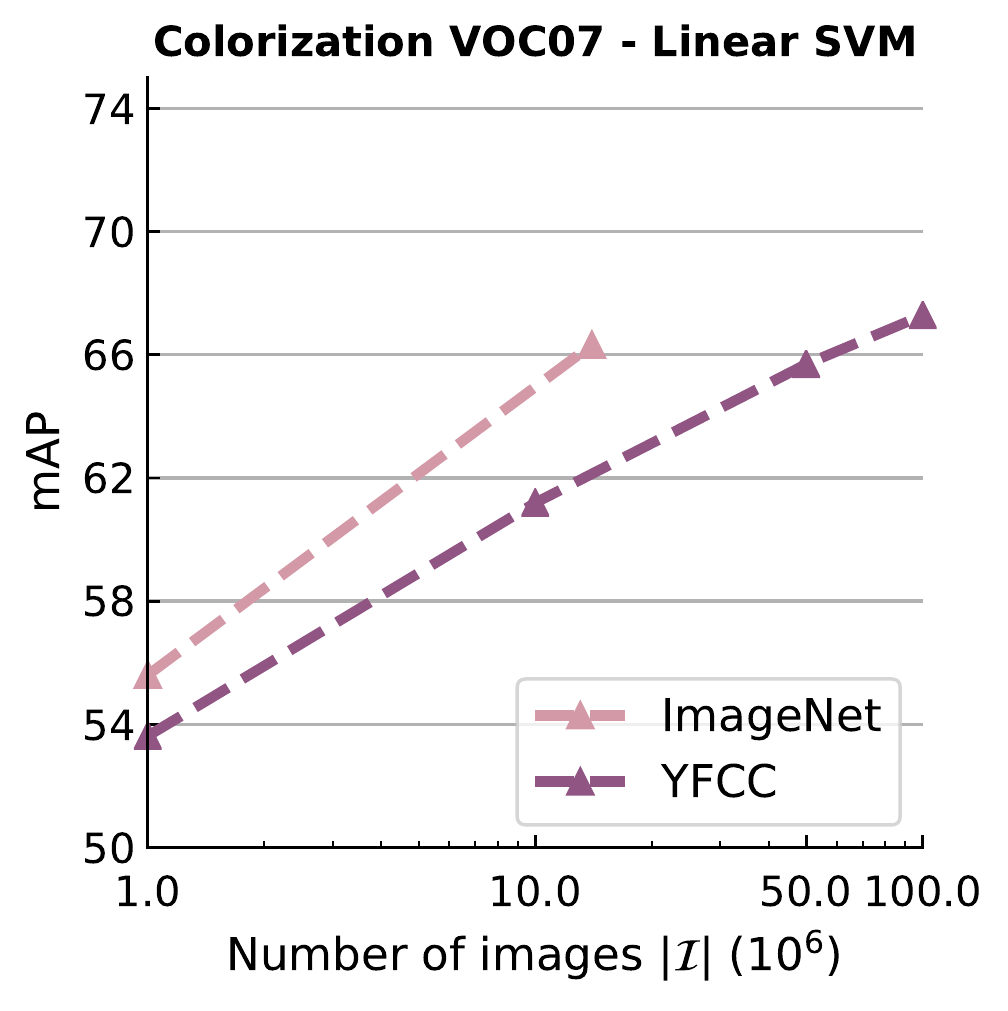} &
    \includegraphics[width=0.247\textwidth]{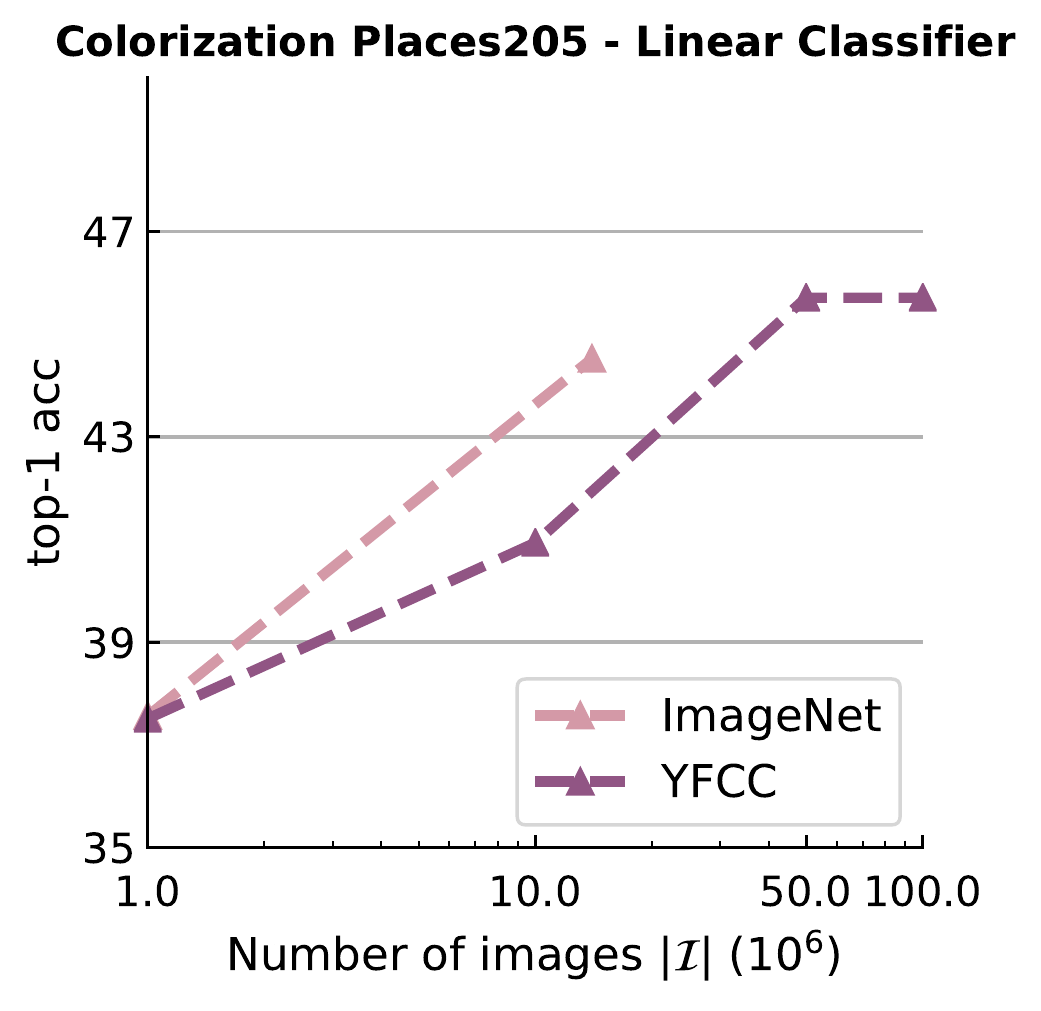}
    \\
    (a) & (b) & (c) & (d)
    \end{tabular}
    \vspace{-0.1in}
    \captionof{figure}{\textbf{Relationship between pre-training and transfer domain:} We vary pre-training data domain - (\ImNetDataset-[1k, 22k], subsets of \YFCCFull) and observe transfer performance on the \VOCseven and \Places classification tasks. The similarity between the pre-training and transfer task domain shows a strong influence on transfer performance.}
    \label{fig:pretrain_domain}
    }
\end{table*}
}

\section{Pre-training and Transfer Domain Relation}
\label{sec:pretrain_domain}
Thus far, we have kept the pre-training dataset and the transfer dataset/task fixed at \YFCC and \VOCseven respectively. We now add the following pre-training and transfer dataset/task to better understand the relationship between pre-training and transfer performance.
\myline
\par \noindent \textbf{Pre-training dataset:} We use both the \ImNetDataset~\cite{deng2009imagenet} and YFCC datasets from \cref{tab:pretrain_datasets}. Although the \ImNetDataset datasets~\cite{ILSVRC15,deng2009imagenet} have supervised labels, we use them (without labels) to study the effect of the pre-training domain.
\myline
\par \noindent \textbf{Transfer dataset and task:} We further evaluate on the \Places scene classification task~\cite{zhou2014learning}. In contrast to the object centric \VOCseven dataset, \Places is a scene centric dataset. Following the investigation setup from \cref{sec:scaling_ablation}, we keep the feature representations of the ConvNets fixed. As the \Places dataset has $>$2M images, we follow~\cite{zhang2017split} and train linear classifiers using SGD. We use a batchsize of $256$, learning rate of $0.01$ decayed by a factor of $10$ after every $40k$ iterations, and train for $140k$ iterations. Full details are provided in the supplementary material.
\myline
\par \noindent \textbf{Observations:} In \cref{fig:pretrain_domain}, we show the results of using different pre-training datasets and transfer datasets/tasks. Comparing Figures~\ref{fig:pretrain_domain} (a) and (b), we make the following observations for the \jigsaw method:
\begin{itemize}[leftmargin=*,noitemsep,nolistsep]
    \item On the \VOCseven classification task, pre-training on \ImNetFull (14M images) transfers as well as pre-training on \YFCCFull (100M images).
    \item However, on the \Places classification task, pre-training on \YFCCone (1M images) transfers as well as pre-training on \ImNetFull (14M images).
\end{itemize}
 
 We note a similar trend for the \colorization problem wherein pre-training \ImNetDataset, rather than \YFCC, provides a greater benefit when transferring to \VOCseven classification (also noted in~\cite{doersch2015unsupervised,caron2018deep,joulin2016learning}). A possible explanation for this benefit is that the domain (image distribution) of \ImNetDataset is closer to \VOCseven (both are object-centric) whereas \YFCC is closer to \Places (both are scene-centric). This motivates us to evaluate self-supervised methods on a variety of different domain/tasks and we propose an extensive evaluation suite next.

\section{Benchmarking Suite for Self-supervision}
\label{sec:benchmarking}
We evaluate self-supervised learning on a diverse set of \numtasks tasks (see \cref{tab:transfer_datasets}) ranging from semantic classification/detection, scene geometry to visual navigation. We select this benchmark based on the principle that a good representation should \emph{generalize} to many different tasks with \emph{limited} supervision and \emph{limited} fine-tuning. We view self-supervised learning as a way to learn feature representations rather than an `initialization method'~\cite{krahenbuhl2015data} and thus perform limited fine-tuning of the features. We first describe each of these tasks and present our benchmarks.

\myline
\par \noindent \textbf{Consistent Evaluation Setup:} We believe that having a consistent evaluation setup, wherein hyperparameters are set fairly for all methods, is important for easier and meaningful comparisons across self-supervised methods. This is crucial to isolate the improvements due to better representations or better transfer optimization\footnote{We discovered inconsistencies across previous methods (different image crops for evaluation, weights re-scaling, pre-processing, longer fine-tuning schedules \etc) which affects the final performance.}.

\myline
\par \noindent\textbf{Common Setup (Pre-training, Feature Extraction and Transfer):} The common transfer process for the benchmark tasks is as follows:

\begin{itemize}[leftmargin=*,noitemsep,nolistsep]
    \item First, we perform self-supervised \textbf{pre-training} using a self-supervised pretext method (\jigsaw or \colorization) on a pre-training dataset from Table~\ref{tab:pretrain_datasets}.
    \item We \textbf{extract features} from various layers of the network. For \alexnet, we do this after every \conv{} layer; for \resnetfifty, we extract features from the last layer of every residual stage denoted, \eg, \texttt{res1}, \texttt{res2} (notation from \cite{girshick2018detectron}) \etc . For simplicity, we use the term \texttt{layer}.
    \item We then evaluate quality of these features (from different self-supervised approaches) by transfer learning, \ie, benchmarking them on various \textbf{transfer} datasets and tasks that have supervision.
\end{itemize}

We summarize these benchmark tasks in Table~\ref{tab:transfer_datasets} and discuss them in the subsections below. For each subsection, we provide \emph{full details} of the training setup: model architecture, hyperparameters \etc in the supplementary material.

\begin{table}[!]
\centering
\setlength{\tabcolsep}{0.2em}\scalebox{0.7}{
\begin{tabular}{l|ccccc}
 \textbf{Method} & \textbf{\mytexttt{layer1}} & \textbf{\mytexttt{layer2}} & \textbf{\mytexttt{layer3}} & \textbf{\mytexttt{layer4}} & \textbf{\mytexttt{layer5}}\\
\shline
\shline
\resnetfifty \ImNet Supervised & 14.8 & 32.6 & 42.1 & 50.8 & 52.5\\
\resnetfifty \Places Supervised  & 16.7 & 32.3 & 43.2 & 54.7 & 62.3\\
\thinline
\resnetfifty Random  & 12.9 & 16.6 & 15.5 & 11.6 & 9.0\\
\thinline
\resnetfifty (NPID)~\cite{wu2018unsupervised}$^\triangleleft$  & 18.1 & 22.3 & 29.7 & 42.1 & 45.5\\
\thinline
\resnetfifty\ \jigsaw \ImNet & \underline{15.1} & 28.8 & 36.8 & 41.2 & 34.4\\
\resnetfifty\ \jigsaw \ImNetFull & 11.0 & \underline{30.2} & 36.4 & 41.5 & 36.4\\
\resnetfifty\ \jigsaw \YFCCFull & 11.3 & 28.6 & \underline{38.1} & 44.8 & 37.4\\
\resnetfifty\ \mytexttt{Coloriz.} \ImNet & 14.7 & 27.4 & 32.7 & 37.5 & 34.8\\
\resnetfifty\ \mytexttt{Coloriz.} \ImNetFull  & \underline{15.0} & \textbf{30.5} & 37.8 & 44.0 & \textbf{41.5}\\
\resnetfifty\ \mytexttt{Coloriz.} \YFCCFull & \textbf{15.2} & \underline{30.4} & \textbf{38.6} & \textbf{45.4} & \textbf{41.5}\\
\shline
\shline
\end{tabular}}
\vspace{-0.08in}
\caption{\textbf{ResNet-50 top-1 center-crop accuracy for linear classification on \Places dataset} (\cref{sec:image_cls_all}). Numbers with $^\triangleleft$ use a different fine-tuning procedure. All other models follow the setup from Zhang \etal~\cite{zhang2017split}.}
\label{tab:rn50_linear_places205}
\end{table}

\subsection{Task 1: Image Classification}
\label{sec:image_cls_all}
We extract image features from various layers of a self-supervised network and train linear classifiers on these fixed representations. We evaluate performance on the classification task for three datasets: \Places, \VOCseven and \COCO. We report results for \resnetfifty in the main paper; \alexnet results are in the supplementary material.
\myline
\par \noindent \textbf{\Places}: 
We strictly follow the training and evaluation setup from Zhang \etal~\cite{zhang2017split} so that we can draw comparisons to existing works (and re-evaluate the model from~\cite{caron2018deep}). We use a batchsize of $256$, learning rate of $0.01$ decayed by a factor of $10$ after every $40k$ iterations, and train for $140k$ iterations using SGD on the \texttt{train} split. We report the top-1 center-crop accuracy on the \texttt{val} split for \resnetfifty in \cref{tab:rn50_linear_places205} and \alexnet in \cref{tab:alexnet_linear_places205}.

\begin{table}[!t]
\centering
\setlength{\tabcolsep}{0.2em}\scalebox{0.7}{
\begin{tabular}{l|ccccc}
 & \multicolumn{5}{c}{\textbf{\Places}} \\
 \textbf{Method} & \textbf{\mytexttt{layer1}} & \textbf{\mytexttt{layer2}} & \textbf{\mytexttt{layer3}} & \textbf{\mytexttt{layer4}} & \textbf{\mytexttt{layer5}}\\
\shline
\alexnet \ImNet Supervised  & 22.4 & 34.7 & 37.5 & 39.2 & 38.0\\
\alexnet \Places Supervised  & 23.2 & 35.6 & 39.8 & 43.5 & 44.8\\
\thinline
\alexnet Random  & 15.7 & 20.8 & 18.5 & 18.2 & 16.6\\
\thinline
\alexnet (Jigsaw)~\cite{noroozi2016unsupervised} & 19.7 & 26.7 & 31.9 & 32.7 & 30.9\\
\alexnet (Colorization)~\cite{zhang2016colorful} & 16.0 & 25.7 & 29.6 & 30.3 & 29.7\\
\alexnet (SplitBrain)~\cite{zhang2017split} & 21.3 & 30.7 & 34.0 & 34.1 & 32.5\\
\alexnet (Counting)~\cite{noroozi2017representation} & 23.3 & 33.9 & 36.3 & 34.7 & 29.6\\
\alexnet (Rotation)~\cite{gidaris2018unsupervised}$^\triangleleft$ & 21.5 & 31.0 & 35.1 & 34.6 & 33.7\\
\alexnet (DeepCluster)~\cite{caron2018deep} &  17.1 & 28.8 & 35.2 & 36.0 & 32.2 \\
\thinline
\alexnet\ \jigsaw \ImNet & \underline{23.7} & 33.2 & 36.6 & 36.3 & 31.9\\
\alexnet\ \jigsaw \ImNetFull  & \textbf{24.2} & \textbf{34.7} & \underline{37.7} & 37.5 & 31.7\\
\alexnet\ \jigsaw \YFCCFull  & \underline{24.1} & \textbf{34.7} & \textbf{38.1} & \textbf{38.2} & 31.6\\
\alexnet\ \mytexttt{Coloriz.} \ImNet & 18.1 & 28.5 & 30.2 & 31.3 & 30.3\\
\alexnet\ \mytexttt{Coloriz.} \ImNetFull & 18.9 & 30.3 & 33.4 & 34.9 & 34.2\\
\alexnet\ \mytexttt{Coloriz.} \YFCCFull & 18.4 & 30.0 & 33.4 & 34.8 & \textbf{34.6}\\
\shline
\end{tabular}}
 \vspace{-0.1in}
 \caption{\textbf{AlexNet top-1 center-crop accuracy for linear classification on \Places dataset} (\cref{sec:image_cls_all}). Numbers for ~\cite{noroozi2016unsupervised,zhang2016colorful} are from ~\cite{zhang2017split}. Numbers with $^\triangleleft$ use a different fine-tuning schedule.} \label{tab:alexnet_linear_places205}
\end{table}

\myline
\par \noindent \textbf{\VOCseven and \COCO}: 
For smaller datasets that fit in memory, we follow~\cite{owens2016ambient} and train linear SVMs~\cite{boser1992training} on the frozen feature representations using LIBLINEAR package~\cite{lin2008liblinear}. We train on \texttt{trainval} split of \VOCseven dataset, and evaluate on \texttt{test} split of \VOCseven. \cref{tab:rn50_linearsvm_voc2007} shows results on \VOCseven for \resnetfifty. \alexnet and \COCO~\cite{COCO} results are provided in the supplementary material.

\begin{table}[!]
\centering
\setlength{\tabcolsep}{0.2em}\scalebox{0.75}{
\begin{tabular}{l|ccccc}
\textbf{Method} & \textbf{\mytexttt{layer1}} & \textbf{\mytexttt{layer2}} & \textbf{\mytexttt{layer3}} & \textbf{\mytexttt{layer4}} & \textbf{\mytexttt{layer5}}\\
\shline
\shline
\resnetfifty \ImNet Supervised & 24.5 & 47.8 & 60.5 & 80.4 & 88.0\\
\resnetfifty \Places Supervised & 28.2 & 46.9 & 59.1 & 77.3 & 80.8\\
\thinline
\resnetfifty Random & 9.6 & 8.3 & 8.1 & 8.0 & 7.7\\
\thinline
\resnetfifty\ \jigsaw \ImNet & \textbf{27.1} & 45.7 & 56.6 & 64.5 & 57.2\\
\resnetfifty\ \jigsaw \ImNetFull & 20.2 & \textbf{47.7} & 57.7 & \textbf{71.9} & \textbf{64.8}\\
\resnetfifty\ \jigsaw \YFCCFull & 20.4 & 47.1 & \textbf{58.4} & 71.0 & 62.5\\
\resnetfifty\ \mytexttt{Coloriz.} \ImNet & 24.3 & 40.7 & 48.1 & 55.6 & 52.3\\
\resnetfifty\ \mytexttt{Coloriz.} \ImNetFull & 25.8 & 43.1 & 53.6 & 66.1 & 62.7\\
\resnetfifty\ \mytexttt{Coloriz.} \YFCCFull & 26.1 & 42.3 & 53.8 & 67.2 & 61.4\\
\shline
\shline
\end{tabular}}
\vspace{-0.05in}
\caption{\textbf{\resnetfifty Linear SVMs} mAP on \VOCseven classification (\cref{sec:image_cls_all}).}
\label{tab:rn50_linearsvm_voc2007}
\end{table}

\myline
\par \noindent \textbf{Observations}: We see a significant accuracy gap between self-supervised and supervised methods despite our scaling efforts. This is expected as unlike self-supervised methods, both the supervised pre-training and benchmark transfer tasks solve a semantic image classification problem.

\subsection{Task 2: Low-shot Image Classification}
\label{sec:lowshot}
It is often argued that a good representation should not require many examples to learn about a concept. Thus, following~\cite{wang2016learning}, we explore the quality of feature representation when per-category examples are few (unlike \cref{sec:image_cls_all}).

\par \noindent \textbf{Setup:} We vary the number $k$ of positive examples (per class) and use the setup from \cref{sec:image_cls_all} to train linear SVMs on \Places and \VOCseven datasets. We perform this evaluation for \resnetfifty only. For each combination of $k$/dataset/method, we report the mean and standard deviation of 5 independent samples of the training data evaluated on a fixed test set (\texttt{test} split for \VOCseven and \texttt{val} split for \Places). We show results for the \jigsaw method in \cref{fig:lowshot_voc}; \colorization results are in the supplementary material as we draw the same observations.

\begin{figure}
    \centering
    \includegraphics[width=0.5\textwidth]{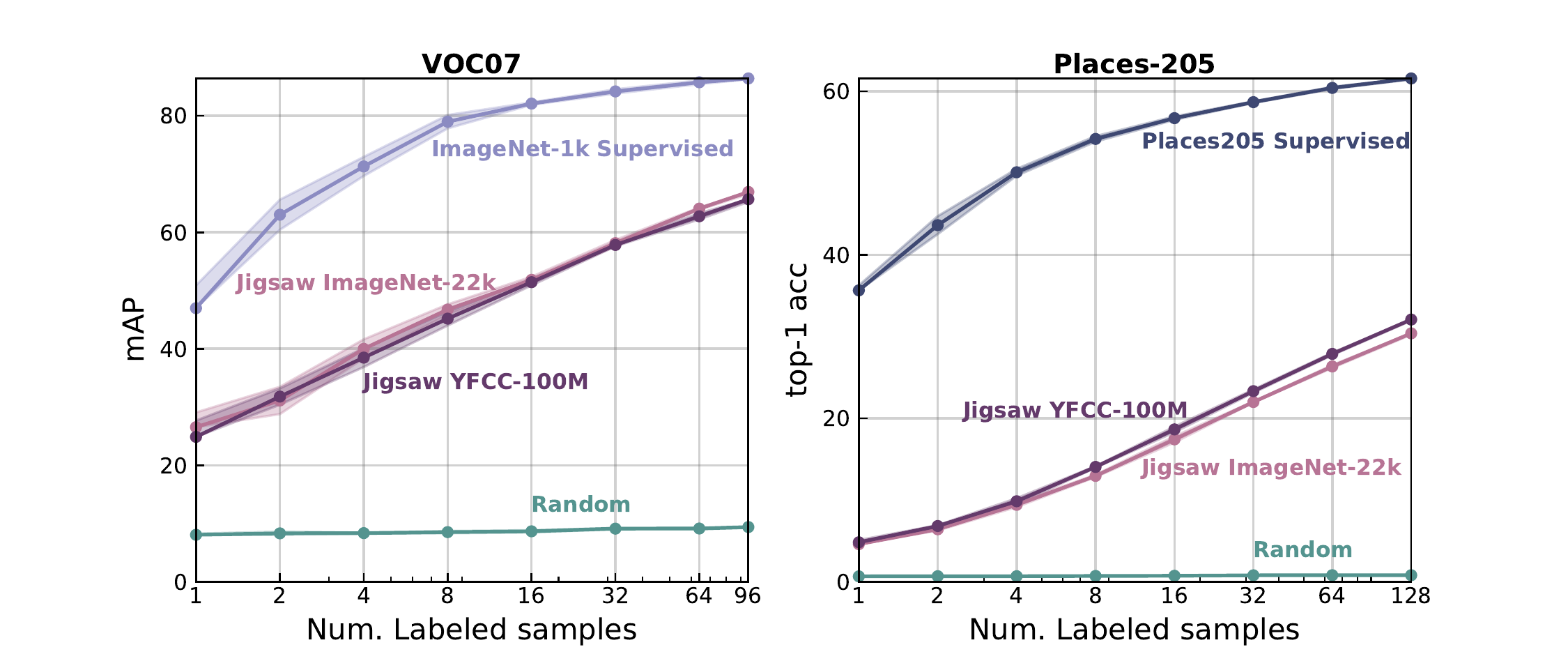}
        \caption{\textbf{Low-shot Image Classification} on the \VOCseven and \Places datasets using linear SVMs trained on the features from the best performing layer for \resnetfifty. We vary the number of labeled examples (per class) used to train the classifier and report the performance on the \texttt{test} set. We show the mean and standard deviation across five runs (\cref{sec:lowshot}).}
    \label{fig:lowshot_voc}
\end{figure}

\myline
\par \noindent \textbf{Observations:} We report results for the best performing layer \texttt{res4} (notation from~\cite{girshick2018detectron}) for \resnetfifty on \VOCseven and \Places in \cref{fig:lowshot_voc}. In the supplementary material, we show that for the lower layers, similar to \cref{tab:rn50_linear_places205}, the self-supervised features are competitive to their supervised counterpart in low-shot setting on both the datasets. However, for both \VOCseven and \Places, we observe a significant gap between supervised and self-supervised settings on their `best' performing layer. This gap is much larger at lower sample size, \eg, at $k\!=\!1$ it is 30 points for \Places, whereas at higher values (full-shot in \cref{tab:rn50_linear_places205}) it is 20 points.

\subsection{Task 3: Visual Navigation}
\label{sec:navigation}
In this task, an agent receives a stream of images as input and learns to navigate to a pre-defined location to get a reward. The agent is spawned at random locations and must build a contextual map in order to be successful at the task. 
\myline
\par \noindent \textbf{Setup:} We use the setup from~\cite{sax2018mid} who train an agent using reinforcement learning (PPO~\cite{schulman2017proximal}) in the Gibson environment~\cite{xia2018gibson}. The agent uses \emph{fixed} feature representations from a ConvNet for this task and only updates the policy network. We evaluate the representation of layers \texttt{res3}, \texttt{res4}, \texttt{res5} (notation from ~\cite{girshick2018detectron}) of a \resnetfifty by separately training agents for these settings. We use the training hyperparameters from~\cite{sax2018mid}, who use a rollout of size 512 and optimize using Adam~\cite{kingma2014adam}.

\myline
\par
\begin{figure*}[!t]
    \centering
    \includegraphics[width=0.9\textwidth]{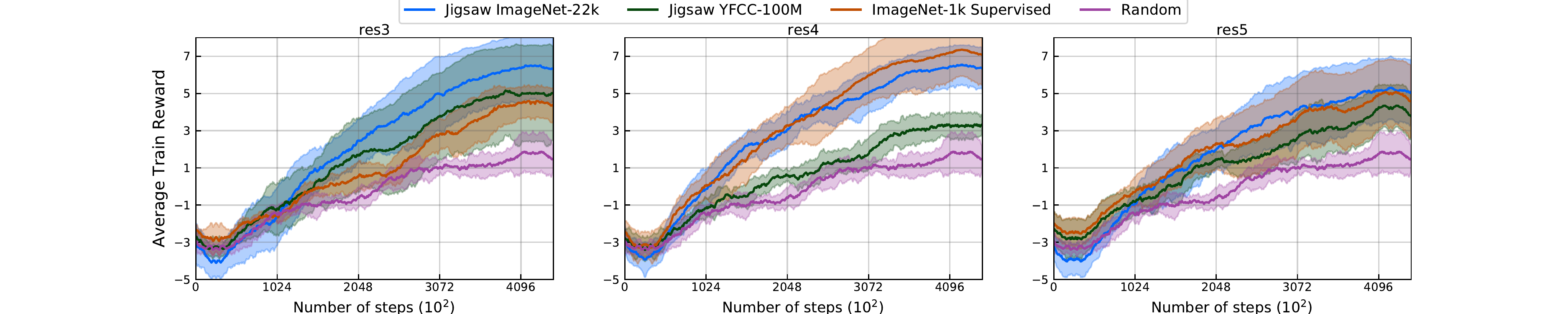}
    \vspace{-0.1in}
    \caption{\textbf{Visual Navigation.} We train an agent on the navigation task in the Gibson environment. The agent is trained using reinforcement learning and uses fixed ConvNet features. We show results for different layers features of \resnetfifty trained on both supervised and self-supervised settings (\cref{sec:navigation}).}
    \label{fig:gibson}
\end{figure*}

 \noindent \textbf{Observations:} \cref{fig:gibson} shows the average training reward (and variance) across 5 runs. Using the \texttt{res3} layer features, we observe that our \jigsaw\ \ImNetDataset model gives a much \textbf{higher training reward} and is more \textbf{sample efficient} (higher reward with fewer steps) than its supervised counterpart. The deeper \texttt{res4} and \texttt{res5} features perform similarly for the supervised and self-supervised networks. We also observe that self-supervised pre-training on the \ImNetDataset domain outperforms pre-training on the \YFCC domain.

\subsection{Task 4: Object Detection}
\label{sec:object_detection}

\begin{table}[!]
\centering
\setlength{\tabcolsep}{0.7em}\scalebox{0.80}{
\begin{tabular}{l|cc}
\textbf{Method} & \textbf{\VOCseven} & \textbf{\VOCseventwelve}\\
\shline
\resnetfifty \ImNet Supervised$^*$ & 66.7 $\pm$ 0.2 & 71.4 $\pm$ 0.1\\
\resnetfifty \ImNet Supervised & \textbf{68.5} $\pm$ 0.3 & \textbf{75.8} $\pm$ 0.2\\
\resnetfifty \Places Supervised & 65.3 $\pm$ 0.3 & 73.1 $\pm$ 0.3\\
\thinline
\resnetfifty\ \jigsaw \ImNet  & 56.6 $\pm$ 0.5 & 64.7 $\pm$ 0.2\\
\resnetfifty\ \jigsaw \ImNetFull  & \textbf{67.1} $\pm$ 0.3 & \textbf{73.0} $\pm$ 0.2\\
\resnetfifty\ \jigsaw \YFCCFull  & 62.3 $\pm$ 0.2 & 69.7 $\pm$ 0.1\\
\shline
\end{tabular}}
\vspace{-0.08in}
\caption{\textbf{Detection mAP for frozen \texttt{conv} body} on \VOCseven and \VOCseventwelve using Fast R-CNN with ResNet-50-C4 (mean and std computed over 5 trials). We freeze the \texttt{conv} body for all models. Numbers with $^*$ use \detectron~\cite{girshick2018detectron} default training schedule. All other models use slightly longer training schedule (see \cref{sec:object_detection}).}
\label{tab:detection-Fbody}
\end{table}
\par \noindent \textbf{Setup:} We use the \detectron~\cite{girshick2018detectron} framework to train the Fast R-CNN~\cite{girshick2015fast} object detection model using Selective Search~\cite{uijlings2013selective} object proposals on the \VOCseven and \VOCseventwelve~\cite{Everingham15} datasets. We provide results for Faster R-CNN~\cite{ren2015faster} in the supplementary material. We note that we use the \emph{same training schedule} for both the supervised and self-supervised methods since it impacts final object detection performance significantly. We report mean and standard deviation result of 5 independent runs for \resnetfifty only as \detectron does not support \alexnet.

We freeze the full \texttt{conv} body of Fast R-CNN and only train the RoI heads (last \resnetfifty stage \texttt{res5} onwards). We follow the same setup as in \detectron and only change the training schedule to be slightly longer. Specifically, we train on 2 GPUS for $22k/8k$ schedule on \VOCseven and for $66k/14k$ schedule on \VOCseventwelve (compared to original $15k/5k$ schedule on \VOCseven and $40k/15k$ schedule on \VOCseventwelve). This change improves object detection performance for both supervised and self-supervised methods. 

\myline
\par \noindent \textbf{Observations:} We report results in \cref{tab:detection-Fbody} and note that the self-supervised initialization is competitive with the ImageNet pre-trained initialization on \VOCseven dataset even when fewer parameters are fine-tuned on the detection task. We also highlight that the performance gap between supervised and self-supervised initialization is very low.

\subsection{Task 5: Surface Normal Estimation}
\label{sec:surface_normals}
\par \noindent \textbf{Setup:} We use the surface normal estimation task~\cite{fouhey2013data}, with the evaluation, and dataset splits as formulated in~\cite{wang2015designing,misra2016cross,bansal2016marr}. We use the \NYU~\cite{silberman2012indoor} dataset which consists of indoor scenes and use the surface normals calculated by~\cite{ladicky2014discriminatively}. We use the state-of-the-art PSPNet~\cite{zhao2017pyramid} architecture (implementation~\cite{zhou2018semantic}). This provides a much stronger baseline (our scratch model outperforms the best numbers reported in~\cite{wang2017transitive}). We fine-tune {\tt res5} onwards and train
all the models with the same hyperparameters for 150 epochs. The scratch model (initialized randomly) is trained for 400 epochs. We use the training hyperparameters from~\cite{zhou2018semantic}, \ie, batchsize of 16, learning rate of $0.02$ decayed polynomially with a power of $0.9$ and optimize using SGD.

\myline
\par \noindent \textbf{Observations:} We report the best \texttt{test} set performance for \jigsaw in \cref{tab:surface_normal_frozen} and results for \colorization are provided in the supplementary material. We use the metrics from~\cite{fouhey2013data} which measure the angular distance (error) of the prediction as well as the percentage of pixels within $t^\circ$ of the ground truth. We note that our \jigsaw \YFCCFull self-supervised model \textbf{outperforms both} the supervised models (\ImNet and \Places supervised) across all the metrics by a significant margin, \eg, a \textbf{5 point} gain compared to the \Places supervised model on the number of pixels within $t^\circ\!=\!11.5$ metric. We, thus, conclude that self-supervised methods provide better features compared to supervised methods for 3D geometric tasks.

\begin{table}[!h]
\setlength{\tabcolsep}{0.12em}
\centering
\scalebox{0.75}{
    \begin{tabular}{@{}l|cccccc@{}}
    & \multicolumn{2}{c}{\textbf{Angle Distance}} && \multicolumn{3}{c}{\textbf{Within $t^\circ$}} \\
    \textbf{Initialization} & Mean & Median && 11.25 & 22.5 & 30 \\
    & \multicolumn{2}{c}{\textbf{(Lower is better)}} && \multicolumn{3}{c}{\textbf{(Higher is better)}} \\
    \shline
    \resnetfifty\ \ImNet supervised & 26.4 &	17.1 && 36.1 & 59.2 & 68.5 \\
    \resnetfifty\ \Places supervised & 23.3 & 14.2 && 41.8 & 65.2 & 73.6 \\
     \thinline
    \resnetfifty\ Scratch & 26.3	& 16.1 &&	37.9 &	60.6 &	69.0 \\
    \thinline
    \resnetfifty\ \jigsaw \ImNet & 24.2 & 14.5 && 41.2 & 64.2 & 72.5 \\
    \resnetfifty\ \jigsaw \ImNetFull & 22.6 & 13.4 && 43.7 & 66.8 & 74.7 \\
    \resnetfifty\ \jigsaw \YFCCFull & \textbf{22.4} & \textbf{13.1} && \textbf{44.6} & \textbf{67.4} & \textbf{75.1} \\
                            \shline
    \end{tabular}}
\vspace{-0.08in}
\caption{\textbf{Surface Normal Estimation on the \NYU dataset}. We train \resnetfifty from \texttt{res5} onwards and freeze the \texttt{conv} body below (\cref{sec:surface_normals}).}
\label{tab:surface_normal_frozen}

\end{table}

\section{Legacy Tasks and Datasets}
\label{sec:legacy}
For completeness, we also report results on the evaluation tasks used by previous works. As we explain next, we do not include these tasks in our benchmark suite (\cref{sec:benchmarking}).
\begin{table}[!]
\centering
\setlength{\tabcolsep}{0.7em}\scalebox{0.7}{
\begin{tabular}{l|cc}
\textbf{Method} & \textbf{\VOCseven} & \textbf{\VOCseventwelve}\\
\shline
\resnetfifty \ImNet Supervised$^*$ & 69.1 $\pm$ 0.4 & \underline{76.2} $\pm$ 0.4\\
\resnetfifty \ImNet Supervised & \textbf{70.5} $\pm$ 0.4 & \textbf{76.2} $\pm$ 0.1\\
\resnetfifty \Places Supervised & 67.2 $\pm$ 0.2 & 74.5 $\pm$ 0.4\\
\thinline
\resnetfifty\ \jigsaw \ImNet  & 61.4 $\pm$ 0.2 & 68.3 $\pm$ 0.4\\
\resnetfifty\ \jigsaw \ImNetFull  & \textbf{69.2} $\pm$ 0.3 & \textbf{75.4} $\pm$ 0.2\\
\resnetfifty\ \jigsaw \YFCCFull  & 66.6 $\pm$ 0.1 & 73.3 $\pm$ 0.4\\
\shline
\end{tabular}}
\vspace{-0.1in}
\caption{\textbf{Detection mAP for full fine-tuning} on \VOCseven and \VOCseventwelve using Fast R-CNN with ResNet-50-C4 (mean and std computed over 5 trials) (\cref{sec:legacy}). Numbers with $^*$ use \detectron~\cite{girshick2018detectron} default training schedule. All other models use a slightly longer training schedule.}
\label{tab:detection-fullft}
\end{table}

\myline
\par \noindent {\bf Full fine-tuning for transfer learning:} This setup fine-tunes all parameters of a self-supervised network and views it as an initialization method. We argue that this view evaluates not only the quality of the representations but also the initialization and optimization method. For completeness, we report results for \alexnet and \resnetfifty on \VOCseven classification in the supplementary material.

\myline
\par \noindent {\bf VOC07 Object Detection with Full Fine-tuning:} This task fine-tunes \emph{all} the weights of a network for the object detection task. We use the same settings as in \cref{sec:object_detection} and report results for supervised and \jigsaw self-supervised methods in Table~\ref{tab:detection-fullft}. Without any bells and whistles, our self-supervised model initialization \textbf{matches the performance} of the supervised initialization on both \VOCseven and \VOCseventwelve. We note that self-supervised pre-training on \ImNetDataset performs better than \YFCC (similar to \cref{sec:pretrain_domain}).

\myline
\par \noindent {\bf \ImNetDataset Classification using Linear Classifiers:} While the task itself is meaningful, we do not include it in our benchmark suite for two reasons:
\begin{enumerate}[noitemsep,leftmargin=*,nolistsep]
    \item For supervised representations, the widely used baseline is trained on \ImNet dataset. Hence, evaluating also on the same dataset (\ImNet) does \emph{not} test generalization of the supervised baseline.
    \item Most existing self-supervised approaches~\cite{zhang2017split,doersch2015unsupervised} use \ImNet for pre-training and evaluate the representations on the same dataset. As observed in \cref{sec:pretrain_domain}, pre-training and evaluating in the \emph{same} domain biases evaluation. Further, the bias is accentuated as we pre-train the self-supervised features and learn the linear classifiers (for transfer) on \emph{identical} images.
\end{enumerate}

\noindent To compare with existing methods, we report results on \ImNet classification for \alexnet in Table~\ref{tab:alexnet_linear_in1k} (setup from \cref{sec:image_cls_all}). We report results on \resnetfifty in the supplementary material.
\begin{table}[!t]
\centering
\setlength{\tabcolsep}{0.2em}\scalebox{0.7}{
\begin{tabular}{l|ccccc}
 & \multicolumn{5}{c}{\textbf{\ImNet}}\\
 \textbf{Method} & \textbf{\mytexttt{layer1}} & \textbf{\mytexttt{layer2}} & \textbf{\mytexttt{layer3}} & \textbf{\mytexttt{layer4}} & \textbf{\mytexttt{layer5}}\\
\shline
\alexnet \ImNet Supervised & 19.4 & 37.1 & 42.5 & 48.0 & 49.6 \\
\alexnet \Places Supervised & 18.9 & 35.5 & 38.9 & 40.9 & 37.3\\
\thinline
\alexnet Random & 11.9 & 17.2 & 15.2 & 14.8 & 13.5\\
\thinline
\alexnet (Jigsaw)~\cite{noroozi2016unsupervised} & 16.2 & 23.3 & 30.2 & 31.7 & 29.6  \\
\alexnet (Colorization)~\cite{zhang2016colorful} & 13.1 & 24.8 & 31.0 & 32.6 & 31.8\\
\alexnet (SplitBrain)~\cite{zhang2017split} & 17.7 & 29.3 & 35.4 & 35.2 & 32.8 \\
\alexnet (Counting)~\cite{noroozi2017representation} & \textbf{23.3} & \textbf{33.9} & 36.3 & 34.7 & 29.6\\
\alexnet (Rotation)~\cite{gidaris2018unsupervised}$^\triangleleft$ & 18.8 & 31.7 & 38.7 & 38.2 & 36.5\\

\alexnet (DeepCluster)~\cite{caron2018deep} & 13.4 & 28.5 & 37.4 & \textbf{39.2} & 35.7 \\
\thinline
\alexnet\ \jigsaw \ImNet & 20.2 & 32.9 & 36.5 & 36.1 & 29.2\\
\alexnet\ \jigsaw \ImNetFull & 20.2 & \textbf{33.9} & \textbf{38.7} & 37.9 & 27.5\\
\alexnet\ \jigsaw \YFCCFull & 20.2 & \underline{33.4} & 38.1 & 37.4 & 25.8\\
\alexnet\ \mytexttt{Coloriz.} \ImNet & 14.1 & 27.5 & 30.6 & 32.1 & 31.1\\
\alexnet\ \mytexttt{Coloriz.} \ImNetFull & 15.0 & 30.5 & 35.5 & 37.9 & \textbf{37.4}\\
\alexnet\ \mytexttt{Coloriz.} \YFCCFull & 14.4 & 28.8 & 33.2 & 35.3 & 34.0\\
\shline
\end{tabular}}
 \vspace{-0.1in}
 \caption{\textbf{AlexNet top-1 center-crop accuracy for linear classification on \ImNet}. Numbers for ~\cite{noroozi2016unsupervised,zhang2016colorful} are from ~\cite{zhang2017split}. Numbers with $^\triangleleft$ use a different fine-tuning schedule.}
\label{tab:alexnet_linear_in1k}
\end{table}

\section{Conclusion}
\vspace{-0.08in}
In this work, we studied the effect of scaling two self-supervised approaches along three axes: data size, model capacity and problem complexity. Our results indicate that transfer performance increases log-linearly with the data size. The quality of the representations also improves with higher capacity models and problem complexity. More interestingly, we observe that the performance improvements on the the three axes are complementary (\cref{sec:scaling_ablation}). We obtain \textbf{state-of-the-art} results on linear classification using the \ImNet and \Places datasets. We also propose a benchmark suite of \numtasks diverse tasks to evaluate the quality of our learned representations.
Our self-supervised learned representation: (a) \textbf{outperforms supervised} baseline on task of surface normal estimation; (b) performs competitively (or better) compared to supervised-ImageNet baseline on navigation task; (c) \textbf{matches the supervised object detection} baseline even with little fine-tuning; (d) performs worse than supervised counterpart on task of image classification and low-shot classification. We believe future work should focus on designing tasks that are complex enough to exploit large scale data and increased model capacity. Our experiments suggest that scaling self-supervision is crucial but there is still a long way to go before definitively surpassing supervised pre-training.

\myline
{\footnotesize{\par \noindent \textbf{Acknowledgements:} We would like to thank Richard Zhang, Mehdi Noroozi, and Andrew Owens for helping understand the experimental setup in their respective works. Rob Fergus and L{\'e}on Bottou for helpful discussions and valuable feedback. Alexander Sax, Bradley Emi, and Saurabh Gupta for helping with the Gibson experiments; Aayush Bansal and Xiaolong Wang for their help in the surface normal experiments. Ross Girshick and Piotr Doll{\'a}r for helpful comments on the manuscript.}}

{\small
\bibliographystyle{ieee}
\bibliography{refs}
}

\newpage
\appendix
\section*{Supplementary Material}
The supplementary material is organized as follows
\begin{itemize}
    \item Section~\ref{sec:model_arch_pretext} provides architecture details for all the self-supervised networks.
    \item Section~\ref{sec:model_arch_transfer} provides architecture details for all the transfer tasks.
    \item Section~\ref{sec:hyperparam_pretrain} lists the hyperparameters for the self-supervised pre-training step.
    \item Section~\ref{sec:hyperparam_benchmark} lists the hyperparameters for the benchmark tasks used in Section 6 of the main paper.
    \item Section~\ref{sec:hyperparam_legacy} lists the hyperparameters for the legacy tasks used in Section 7 of the main paper.
    \item Section~\ref{sec:extra_problem_complexity} shows results using additional ways of increasing problem complexity for the self-supervised tasks.
    \item Section~\ref{sec:extra_results} shows additional results on object detection, surface normal estimation and image classification.
\end{itemize}

\section{Model architectures for pretext tasks}
\label{sec:model_arch_pretext}
We describe the exact architecture we use for pre-training on \jigsaw and \colorization pretext tasks below. 

\subsection{\alexnet \jigsawH Pretext}
We describe the \alexnet architecture used for \jigsaw model training. We use the same architecture as ~\cite{noroozi2016unsupervised}. Full details in Table~\ref{tab:alexnet_jigsaw_pretext}. 
\begin{table*}[!b]
\centering
\setlength{\tabcolsep}{0.9em}\scalebox{0.9}{
\begin{tabular}{l|ccccccc}
 \textbf{Layer} & \textbf{\mytexttt{X}} & \textbf{\mytexttt{C}} & \textbf{\mytexttt{K}} & \textbf{\mytexttt{S}} & \textbf{\mytexttt{P}} & \textbf{\mytexttt{G}} & \textbf{\mytexttt{D}}\\
\shline
\shline
\textbf{\mytexttt{data}} & 64    & 27& -- & -- & -- & -- & --\\
\textbf{\mytexttt{data\_{split}}} & 64   & 3 & -- & -- & -- & -- & --\\
\textbf{\mytexttt{conv1}} & 27 & 96   & 11& 2 & 0 & 1 & 1\\
\textbf{\mytexttt{pool1}} & 13 & 96   & 3 & 2 & 0 & -- & 1\\
\textbf{\mytexttt{conv2}} & 13 & 256  & 5 & 1 & 2 & 2 & 1\\
\textbf{\mytexttt{pool2}} & 6  & 256  & 3 & 2 & 0 & -- & 1\\
\textbf{\mytexttt{conv3}} & 6  & 384  & 3 & 1 & 1 & 1 & 1\\
\textbf{\mytexttt{conv4}} & 6  & 384  & 3 & 1 & 1 & 2 & 1\\
\textbf{\mytexttt{conv5}} & 6  & 256  & 3 & 1 & 1 & 2 & 1\\
\textbf{\mytexttt{pool5}} & 2  & 256  & 3 & 2 & 0 & -- & 1\\
\textbf{\mytexttt{fc6}} & 1    & 1024 & 1 & 1 & 0 & -- & 2\\
\textbf{\mytexttt{concat}} & 1 & 9216 & 1 & 1 & 0 & -- & 1\\
\textbf{\mytexttt{fc7}} & 1    & 4096 & 1 & 1 & 0 & -- & 1\\
\textbf{\mytexttt{fc8}} & 1    & $^*$ & 1 & 1 & 0 & -- & 1\\

\shline
\shline
\end{tabular}}

\caption{\textbf{AlexNet architecture used for \jigsaw pretext task}. \textbf{X} spatial resolution of layer, \textbf{C} number
of channels in layer; \textbf{K} conv or pool kernel size; \textbf{S} computation stride; \textbf{D} kernel dilation; \textbf{P} padding; \textbf{G} group convolution, last layer is removed during transfer evaluation. Number with * depends on the size per permutation set used to train jigsaw puzzle.}
\label{tab:alexnet_jigsaw_pretext}
\end{table*}

\subsection{\alexnet \colorizationH Pretext}
We use the same architecture setup as ~\cite{zhang2016colorful} and recommend the reader to consult their implementation. Every \texttt{conv} layer is followed by \spatialbn $+$ \relu combination. Full details in Table~\ref{tab:alexnet_arch_color_pretext}.
\begin{table*}[!b]
\centering
\setlength{\tabcolsep}{0.9em}\scalebox{0.9}{
\begin{tabular}{l|ccccccc}
 \textbf{Layer} & \textbf{\mytexttt{X}} & \textbf{\mytexttt{C}} & \textbf{\mytexttt{K}} & \textbf{\mytexttt{S}} & \textbf{\mytexttt{P}} & \textbf{\mytexttt{G}} & \textbf{\mytexttt{D}}\\
\shline
\shline
\textbf{\mytexttt{data}} & 180 & 1 & -- & -- & -- & -- & --\\
\textbf{\mytexttt{conv1}} & 45 & 96 & 11 & 4 & 5 & 1 & 1\\
\textbf{\mytexttt{pool1}} & 23 & 96 & 3 & 2 & 1 & -- & 1\\
\textbf{\mytexttt{conv2}} & 23 & 256 & 5 & 1 & 2 & 2 & 1\\
\textbf{\mytexttt{pool2}} & 12 & 256 & 3 & 2 & 1 & -- & 1\\
\textbf{\mytexttt{conv3}} & 12 & 384 & 3 & 1 & 1 & 1 & 1\\
\textbf{\mytexttt{conv4}} & 12 & 384 & 3 & 1 & 1 & 2 & 1\\
\textbf{\mytexttt{conv5}} & 12 & 256 & 3 & 1 & 1 & 2 & 1\\
\textbf{\mytexttt{pool5}} & 12 & 256 & 3 & 1 & 1 & -- & 1\\
\textbf{\mytexttt{fc6}} & 12 & 4096 & 1 & 1 & 0 & -- & 2\\
\textbf{\mytexttt{fc7}} & 12 & 4096 & 1 & 1 & 0 & -- & 1\\
\textbf{\mytexttt{fc8}} & 12 & $^*$ & 1 & 1 & 0 & -- & 1\\
\shline
\shline
\end{tabular}}
\caption{\textbf{AlexNet architecture used for \colorization pretext task}. \textbf{X} spatial resolution of layer, \textbf{C} number
of channels in layer; \textbf{K} conv or pool kernel size; \textbf{S} computation stride; \textbf{D} kernel dilation; \textbf{P} padding; \textbf{G} group convolution, last layer is removed during transfer evaluation. Number with * depends on the colorization bin size.}
\label{tab:alexnet_arch_color_pretext}
\end{table*}

\subsection{\alexnet Supervised }
We follow the CaffeNet BVLC exact architecture and directly use the pre-trained model weights. We refer reader to~\cite{jia2014caffe} for exact architecture details. We did not train our \alexnet supervised model to avoid introducing any differences in results. 

\subsection{\resnetfifty \jigsawH Pretext}
\label{sec:rn50_jigsaw_pretext}
The \resnetfifty architecture used to train \jigsaw model is described below. The jigsaw model is trained using $N$-way Siamese ConvNet with shared weights. We describe the one siamese branch only in Table~\ref{tab:rn50_jigsaw_pretext}. Also note that, after the $N$-way siamese branches are concatenated, we have single branch left.
\begin{table*}[!b]
\centering
\setlength{\tabcolsep}{0.9em}\scalebox{0.9}{
\begin{tabular}{l|ccccccc}
 \textbf{Layer} & \textbf{\mytexttt{X}} & \textbf{\mytexttt{C}} & \textbf{\mytexttt{K}} & \textbf{\mytexttt{S}} & \textbf{\mytexttt{P}} & \textbf{\mytexttt{G}}\\
\shline
\shline
\textbf{\mytexttt{data}} & 64 & 27 & -- & -- & -- & --\\
\textbf{\mytexttt{data\_{split}}} & 64 & 3 & -- & -- & -- & --\\
\textbf{\mytexttt{conv1}} & 32 & 64 & 7 & 2 & 3 & 1\\
\textbf{\mytexttt{maxpool}} & 16 & 64 & 3 & 2 & 1 & --\\
\textbf{\mytexttt{\texttt{res2}}} & 16 & 256 & $^*$ & $^*$ & $^*$ & 1\\
\textbf{\mytexttt{\texttt{res3}}} & 8 & 512 & $^*$ & $^*$ & $^*$ & 1\\
\textbf{\mytexttt{\texttt{res4}}} & 4 & 1024 & $^*$ & $^*$ & $^*$ & 1\\
\textbf{\mytexttt{\texttt{res5}}} & 2 & 2048 & $^*$ & $^*$ & $^*$ & 1\\
\textbf{\mytexttt{avgpool}} & 1 & 2048 & 2 & 1 & 0 & --\\
\textbf{\mytexttt{fc1$^\dagger$}} & 1 & 1024 & 1 & 1 & 0 & 1\\
\textbf{\mytexttt{concat}} & 1 & 9216 & -- & -- & -- & --\\
\textbf{\mytexttt{fc2}} & 1 & $^\triangleleft$ & 1 & 1 & 0 & 1\\
\shline
\shline
\end{tabular}}
\vspace{-0.1in}
\caption{\textbf{\resnetfifty architecture used for \jigsaw pretext task}. \textbf{X} spatial resolution of layer, \textbf{C} number
of channels in layer; \textbf{K} conv or pool kernel size; \textbf{S} computation stride; \textbf{D} kernel dilation; \textbf{P} padding; \textbf{G} group convolution. Layers denoted with \texttt{res} prefix represent the bottleneck residual block. Number with * use the original setting as in~\cite{he2016deep}. Layer with $^\dagger$ is implemented as a \texttt{conv} layer. Number with $^\triangleleft$ depend on the size of permutation set used for training \jigsaw model (see Section 4.3 in main paper)}.
\label{tab:rn50_jigsaw_pretext}
\end{table*}

\subsection{\resnetfifty \colorizationH Pretext}
The \resnetfifty architecture used to train \colorization model is described in Table~\ref{tab:rn50_color_pretext}. We closely follow the architecture as in~\ref{sec:rn50_jigsaw_pretext}.
\begin{table*}[!b]
\centering
\setlength{\tabcolsep}{0.9em}\scalebox{0.9}{
\begin{tabular}{l|ccccccc}
 \textbf{Layer} & \textbf{\mytexttt{X}} & \textbf{\mytexttt{C}} & \textbf{\mytexttt{K}} & \textbf{\mytexttt{S}} & \textbf{\mytexttt{P}} & \textbf{\mytexttt{G}}\\
\shline
\shline
\textbf{\mytexttt{data}} & 180 & 1 & -- & -- & -- & --\\
\textbf{\mytexttt{conv1}} & 90 & 64 & 7 & 2 & 3 & 1\\
\textbf{\mytexttt{maxpool}} & 45 & 64 & 3 & 2 & 1 & --\\
\textbf{\mytexttt{\texttt{res2}}} & 45 & 256 & $^*$ & $^*$ & $^*$ & 1\\
\textbf{\mytexttt{\texttt{res3}}} & 23 & 512 & $^*$ & $^*$ & $^*$ & 1\\
\textbf{\mytexttt{\texttt{res4}}} & 12 & 1024 & $^*$ & $^*$ & $^*$ & 1\\
\textbf{\mytexttt{\texttt{res5}}} & 12 & 2048 & $^*$ & $^*$ & $^*$ & 1\\
\textbf{\mytexttt{avgpool}} & 12 & 2048 & 3 & 1 & 1 & --\\
\textbf{\mytexttt{fc1$^\dagger$}} & 12 & 313 & 6 & 1 & 5 & 1\\
\shline
\shline
\end{tabular}}
\vspace{-0.08in}
\caption{\textbf{\resnetfifty architecture used for \colorization pretext task}. \textbf{X} spatial resolution of layer, \textbf{C} number
of channels in layer; \textbf{K} conv or pool kernel size; \textbf{S} computation stride; \textbf{D} kernel dilation; \textbf{P} padding; \textbf{G} group convolution. Layers denoted with \texttt{res} prefix represent the bottleneck residual block. Number with * use the original setting as in~\cite{he2016deep}. Layer with $^\dagger$ is implemented as a \texttt{conv} layer.}
\label{tab:rn50_color_pretext}
\end{table*}

\subsection{\resnetfifty Supervised}
We strictly follow the same \resnet architecture as in ~\cite{goyal2017accurate} and refer the reader to the work.

\section{Model architectures for Transfer tasks}
\label{sec:model_arch_transfer}
In this section, we describe the exact model architecture we use for various evaluation tasks (including benchmark suite as described in Section 6 of the main paper).

\subsection{\alexnet \colorizationH Transfer}
We use the same architecture setup as ~\cite{zhang2016colorful} and recommend the reader to consult their implementation. Every \texttt{conv} layer is followed by \spatialbn $+$ \relu combination. For evaluation, we downsample \texttt{conv} layers so that the resulting feature map has dimension $~9k$. Full details in Table~\ref{tab:alexnet_arch_color_tune}.
\begin{table*}[!b]
\centering
\setlength{\tabcolsep}{0.9em}\scalebox{0.9}{
\begin{tabular}{l|ccccccc|cccc}
 \textbf{Layer} & \textbf{\mytexttt{X}} & \textbf{\mytexttt{C}} & \textbf{\mytexttt{K}} & \textbf{\mytexttt{S}} & \textbf{\mytexttt{P}} & \textbf{\mytexttt{G}} & \textbf{\mytexttt{D}} & \textbf{\mytexttt{$X_{d}$}}  & \textbf{\mytexttt{$K_{d}$}} & \textbf{\mytexttt{$S_{d}$}} & \textbf{\mytexttt{$P_{d}$}}\\
\shline
\shline
\textbf{\mytexttt{data}} & 227 & 1   & -- & -- & -- & -- & --& -- & -- & -- & --\\
\textbf{\mytexttt{conv1}} & 55 & 96  & 11 & 4  & 0  & 1  & 1 & 10 & 19 & 4  & 0 \\
\textbf{\mytexttt{pool1}} & 27 & 96  & 3  & 2  & 0  & -- & 1 & -- & -- & -- & --\\
\textbf{\mytexttt{conv2}} & 27 & 256 & 5  & 1  & 2  & 2  & 1 & 6  & 12 & 3  & 0 \\
\textbf{\mytexttt{pool2}} & 13 & 256 & 3  & 2  & 0  & -- & 1 & -- & -- & -- & --\\
\textbf{\mytexttt{conv3}} & 13 & 384 & 3  & 1  & 1  & 1  & 1 & 5  & 9  & 1  & 0 \\
\textbf{\mytexttt{conv4}} & 13 & 384 & 3  & 1  & 1  & 2  & 1 & 5  & 9  & 1  & 0 \\
\textbf{\mytexttt{conv5}} & 13 & 256 & 3  & 1  & 1  & 2  & 1 & 6  & 8  & 1  & 0 \\
\textbf{\mytexttt{pool5}} & 6 & 256 & 3 & 1 & 1 & -- & 1     & -- & -- & -- & --\\
\textbf{\mytexttt{fc6}} & 1 & 4096 & 1 & 1 & 5 & -- & 2      & -- & -- & -- & --\\
\textbf{\mytexttt{fc7}} & 1 & 4096 & 1 & 1 & 0 & -- & 1      & -- & -- & -- & --\\
\textbf{\mytexttt{fc8}} & 1 & $^*$ & 1 & 1 & 0 & -- & 1      & -- & -- & -- & --\\
\shline
\shline
\end{tabular}}
\caption{\textbf{AlexNet architecture used for \colorization finetuning}. \textbf{X} spatial resolution of layer, \textbf{C} number
of channels in layer; \textbf{K} conv or pool kernel size; \textbf{S} computation stride; \textbf{D} kernel dilation; \textbf{P} padding; \textbf{G} group convolution, last layer is removed during transfer evaluation. Number with * depends on the colorization bin size. For evaluation, we downsample \texttt{conv} layers so that the resulting feature map has dimension $~9k$. \textbf{$X_{d}$} downsampled spatial resolution; \textbf{$K_{d}$} kernel size of downsample \texttt{avgpool} layer; \textbf{$S_{d}$} stride of downsample \texttt{avgpool} layer; \textbf{$P_{d}$} padding of downsample using \texttt{avgpool} layer.}
\label{tab:alexnet_arch_color_tune}
\end{table*}

\subsection{\alexnet \jigsawH Transfer}
For evaluation, we downsample \texttt{conv} layers by applying an \texttt{avgpool} layer so that the resulting feature map has dimension $~9k$. Full details in Table~\ref{tab:alexnet_jigsaw_tune}.
\begin{table*}[!b]
\centering
\setlength{\tabcolsep}{0.9em}\scalebox{0.9}{
\begin{tabular}{l|ccccccc|cccc}
 \textbf{Layer} & \textbf{\mytexttt{X}} & \textbf{\mytexttt{C}} & \textbf{\mytexttt{K}} & \textbf{\mytexttt{S}} & \textbf{\mytexttt{P}} & \textbf{\mytexttt{G}} & \textbf{\mytexttt{D}} & \textbf{\mytexttt{$X_{d}$}}  & \textbf{\mytexttt{$K_{d}$}} & \textbf{\mytexttt{$S_{d}$}} & \textbf{\mytexttt{$P_{d}$}}\\
\shline
\shline
\textbf{\mytexttt{data}} & 227 & 3   & -- & -- & -- & -- & --& -- & -- & -- & --\\
\textbf{\mytexttt{conv1}} & 109& 96  & 11 & 2  & 0  & 1  & 1 & 10 & 28 & 9  & 0 \\
\textbf{\mytexttt{pool1}} & 54 & 96  & 3  & 2  & 0  & -- & 1 & -- & -- & -- & --\\
\textbf{\mytexttt{conv2}} & 54 & 256 & 5  & 1  & 2  & 2  & 1 & 6  & 24 & 6  & 0 \\
\textbf{\mytexttt{pool2}} & 26 & 256 & 3  & 2  & 0  & -- & 1 & -- & -- & -- & --\\
\textbf{\mytexttt{conv3}} & 26 & 384 & 3  & 1  & 1  & 1  & 1 & 5  & 14  & 3  & 0\\
\textbf{\mytexttt{conv4}} & 26 & 384 & 3  & 1  & 1  & 2  & 1 & 5  & 14  & 3  & 0\\
\textbf{\mytexttt{conv5}} & 26 & 256 & 3  & 1  & 1  & 2  & 1 & 6  & 16  & 2  & 0\\
\textbf{\mytexttt{pool5}} & 12 & 256 & 3 & 2 & 0 & -- & 1     & -- & -- & -- & --\\
\textbf{\mytexttt{fc6}} & 1 & 4096 & 1 & 1 & 0 & -- & 1      & -- & -- & -- & --\\
\textbf{\mytexttt{fc7}} & 1 & 4096 & 1 & 1 & 0 & -- & 1      & -- & -- & -- & --\\
\textbf{\mytexttt{fc8}} & 1 & $^*$ & 1 & 1 & 0 & -- & 1      & -- & -- & -- & --\\
\shline
\shline
\end{tabular}}
\caption{\textbf{AlexNet architecture used for \jigsaw finetuning}. \textbf{X} spatial resolution of layer, \textbf{C} number
of channels in layer; \textbf{K} conv or pool kernel size; \textbf{S} computation stride; \textbf{D} kernel dilation; \textbf{P} padding; \textbf{G} group convolution, last layer is removed during transfer evaluation. Number with * depends on the colorization bin size. For evaluation, we downsample \texttt{conv} layers so that the resulting feature map has dimension $~9k$. \textbf{$X_{d}$} downsampled spatial resolution; \textbf{$K_{d}$} kernel size of downsample \texttt{avgpool} layer; \textbf{$S_{d}$} stride of downsample \texttt{avgpool} layer; \textbf{$P_{d}$} padding of downsample \texttt{avgpool} layer.}
\label{tab:alexnet_jigsaw_tune}
\end{table*}

\subsection{\alexnet Supervised Transfer}
We follow the CaffeNet BVLC exact architecture and directly use the pre-trained model weights. We refer reader to~\cite{jia2014caffe} for exact architecture details.

\subsection{\resnetfifty \jigsawH Transfer}
Table~\ref{tab:rn50_jigsaw_tune} shows the exact architecture used.
\begin{table*}[!b]
\centering
\setlength{\tabcolsep}{0.9em}\scalebox{0.9}{
\begin{tabular}{l|ccccccc|cccc}
 \textbf{Layer} & \textbf{\mytexttt{X}} & \textbf{\mytexttt{C}} & \textbf{\mytexttt{K}} & \textbf{\mytexttt{S}} & \textbf{\mytexttt{P}} & \textbf{\mytexttt{G}} & \textbf{\mytexttt{$X_{d}$}}  & \textbf{\mytexttt{$K_{d}$}} & \textbf{\mytexttt{$S_{d}$}} & \textbf{\mytexttt{$P_{d}$}}\\
\shline
\shline
\textbf{\mytexttt{data}} & 224 & 1 & -- & -- & -- & --& -- & -- & -- & --\\
\textbf{\mytexttt{conv1}} & 112 & 64 & 7 & 2 & 3 & 1 & 12 & 10 & 10  & 4\\
\textbf{\mytexttt{maxpool}} & 56 & 64 & 3 & 2 & 1 & --& -- & -- & -- & --\\
\textbf{\mytexttt{\texttt{res2}}} & 56 & 256 & $^*$ & $^*$ & $^*$ & 1 & 6 & 16 & 8  & 0\\
\textbf{\mytexttt{\texttt{res3}}} & 28 & 512 & $^*$ & $^*$ & $^*$ & 1 & 4 & 13 & 5  & 0\\
\textbf{\mytexttt{\texttt{res4}}} & 14 & 1024 & $^*$ & $^*$ & $^*$ & 1 & 3 & 8 & 3  & 0\\
\textbf{\mytexttt{\texttt{res5}}} & 7 & 2048 & $^*$ & $^*$ & $^*$ & 1 & 2 & 6 & 1  & 0\\
\textbf{\mytexttt{avgpool}} & 1 & 2048 & 7 & 1 & 0 & --& -- & -- & -- & --\\
\textbf{\mytexttt{fc1}} & 1 & $^\dagger$ & 1 & 1 & 0 & --& -- & -- & -- & --\\
\shline
\shline
\end{tabular}}
\vspace{-0.1in}
\caption{\textbf{\resnetfifty architecture used for \jigsaw Transfer task}. \textbf{X} spatial resolution of layer, \textbf{C} number
of channels in layer; \textbf{K} conv or pool kernel size; \textbf{S} computation stride; \textbf{D} kernel dilation; \textbf{P} padding; \textbf{G} group convolution. Layers denoted with \texttt{res} prefix represent the bottleneck residual block. Number with * use the original setting as in~\cite{he2016deep}. Layer with $^\dagger$ depend on the number of output classes. For evaluation, we downsample \texttt{conv} layers so that the resulting feature map has dimension $~9k$. \textbf{$X_{d}$} downsampled spatial resolution; \textbf{$K_{d}$} kernel size of downsample \texttt{avgpool} layer; \textbf{$S_{d}$} stride of downsample \texttt{avgpool} layer; \textbf{$P_{d}$} padding of downsample using \texttt{avgpool} layer.}
\label{tab:rn50_jigsaw_tune}
\end{table*}

\subsection{\colorizationH \resnetfifty Transfer}
Table~\ref{tab:rn50_color_tune} shows the exact architecture used.
\begin{table*}[!b]
\centering
\setlength{\tabcolsep}{0.9em}\scalebox{0.9}{
\begin{tabular}{l|ccccccc|cccc}
 \textbf{Layer} & \textbf{\mytexttt{X}} & \textbf{\mytexttt{C}} & \textbf{\mytexttt{K}} & \textbf{\mytexttt{S}} & \textbf{\mytexttt{P}} & \textbf{\mytexttt{G}} & \textbf{\mytexttt{$X_{d}$}}  & \textbf{\mytexttt{$K_{d}$}} & \textbf{\mytexttt{$S_{d}$}} & \textbf{\mytexttt{$P_{d}$}}\\
\shline
\shline
\textbf{\mytexttt{data}} & 224 & 1 & -- & -- & -- & --& -- & -- & -- & --\\
\textbf{\mytexttt{conv1}} & 112 & 64 & 7 & 2 & 3 & 1 & 12 & 10 & 10  & 4\\
\textbf{\mytexttt{maxpool}} & 56 & 64 & 3 & 2 & 1 & --& -- & -- & -- & --\\
\textbf{\mytexttt{\texttt{res2}}} & 56 & 256 & $^*$ & $^*$ & $^*$ & 1 & 6 & 16 & 8  & 0\\
\textbf{\mytexttt{\texttt{res3}}} & 28 & 512 & $^*$ & $^*$ & $^*$ & 1 & 4 & 13 & 5  & 0\\
\textbf{\mytexttt{\texttt{res4}}} & 14 & 1024 & $^*$ & $^*$ & $^*$ & 1 & 3 & 8 & 3  & 0\\
\textbf{\mytexttt{\texttt{res5}}} & 14 & 2048 & $^*$ & $^*$ & $^*$ & 1 & 2 & 12 & 2  & 0\\
\textbf{\mytexttt{avgpool}} & 1 & 2048 & 14 & 1 & 0 & --& -- & -- & -- & --\\
\textbf{\mytexttt{fc1}} & 1 & $^\dagger$ & 1 & 1 & 0 & --& -- & -- & -- & --\\
\shline
\shline
\end{tabular}}
\vspace{-0.08in}
\caption{\textbf{\resnetfifty architecture used for \colorization Transfer task}. \textbf{X} spatial resolution of layer, \textbf{C} number
of channels in layer; \textbf{K} conv or pool kernel size; \textbf{S} computation stride; \textbf{D} kernel dilation; \textbf{P} padding; \textbf{G} group convolution. Layers denoted with \texttt{res} prefix represent the bottleneck residual block. Number with * use the original setting as in~\cite{he2016deep}. Layer with $^\dagger$ depend on the number of output classes. For evaluation, we downsample \texttt{conv} layers so that the resulting feature map has dimension $~9k$. \textbf{$X_{d}$} downsampled spatial resolution; \textbf{$K_{d}$} kernel size of downsample \texttt{avgpool} layer; \textbf{$S_{d}$} stride of downsample \texttt{avgpool} layer; \textbf{$P_{d}$} padding of downsample using \texttt{avgpool} layer.}
\label{tab:rn50_color_tune}
\end{table*}

\subsection{\resnetfifty Supervised Transfer}
We strictly follow the same \resnet architecture as in ~\cite{goyal2017accurate} and refer the reader to the work.

\section{Pre-training Hyperparameters}
\label{sec:hyperparam_pretrain}

In this section, we describe the pre-training hyperparameters used to pre-train self-supervised methods (\jigsaw and \colorization) for both \alexnet and \resnetfifty models. 

\subsection{\alexnet\ \jigsawH}
For training \alexnet on \jigsaw, we follow the hyperparameters setting from ~\cite{noroozi2016unsupervised}. For the jigsaw problem, we read the original image from the data source, scale the shorter side to 256 and randomly crop out $255\times255$ image.  We make the images grayscale randomly with 50$\%$ probability and we apply color projection with 50$\%$ probability (if the image is still colored). We further divide the image into 3x3 grid with each cell of size 85x85. Next, we randomly crop out a patch of size 64x64 from the 85x85 cell. This patch becomes a piece in jigsaw puzzle. Further, following~\cite{noroozi2016unsupervised}, we apply bias decay for the bias parameter of the model and we also do not apply weight decay to the \texttt{scale} and \texttt{bias} parameter of \spatialbn layers. We train the model on 8-gpus, use minibatch size of 256, initial learning rate (LR) of 0.01 with the learning rate dropped by factor of 10 at certain interval. We use momentum of 0.9, weight decay 5e-4 and \spatialbn weight decay 0. We use nesterov SGD for optimization. The number of training iterations depends on the dataset size we are training on. We describe that next for each different dataset used and also corresponding to the \emph{best} models reported in the main paper.

\par \noindent Model training iterations for Scaling Data size analysis (Section 4.1 of main paper):
\begin{itemize}
    \item \ImNet permutation 2k: Train for 70 epochs with LR schedule: $100k/100k/100k/50k$.
    \item \ImNetFull permutation 2k: Train for 100 epochs with LR schedule: $1584343/1584343/1584343/792171$.
    \item \YFCCone permutation 2k: Train for 70 epochs with LR schedule: $100k/100k/100k/50k$. 
    \item \YFCCten permutation 2k: Train for 70 epochs with LR schedule: $781250/781250/781250/390625$.
    \item \YFCCfifty permutation 2k: Train for 10 epochs only with LR schedule: $558036/558036/558036/279017$.
     \item \YFCCFull permutation 2k: Train for 25 epochs with LR schedule: $2790178/2790178/2790178/1395089$.
    
\end{itemize}

\myline
\par \noindent For the best models (Section 6 of main paper), the training schedule is as follows:

\begin{itemize}
    \item \ImNet permutation 2k: Train for 70 epochs with LR schedule: $100k/100k/100k/50k$.
    \item \ImNetFull permutation 2k: Train for 100 epochs with LR schedule: $1584343/1584343/1584343/792171$.
    \item \YFCCFull permutation 2k: Train for 25 epochs with LR schedule: $2790178/2790178/2790178/1395089$.
\end{itemize}

\subsection{\resnetfifty\ \jigsawH}
For training \resnetfifty on \jigsaw, we closely follow the hyperparameters setting from ~\cite{noroozi2016unsupervised}. Specifically, we read the original image from the data source, scale the shorter side to 256 and randomly crop out $255\times255$ image. We make the images grayscale randomly with 50$\%$ probability and we apply color projection with 50$\%$ probability (if the image is still colored). We further divide the image into 3x3 grid with each cell of size 85x85. Next, we randomly crop out a patch of size 64x64 from the 85x85 cell. This patch becomes a piece in jigsaw puzzle. Further, following~\cite{noroozi2016unsupervised}, we apply bias decay for the bias parameter of the model. We train the model on 8-gpus, use minibatch size of 256, initial learning rate (LR) of 0.1 with the learning rate dropped by factor of 10 after certain steps. We use momentum of 0.9, weight decay 1e-4 and \spatialbn weight decay 0. We use Nesterov SGD for optimization. The number of training iterations depends on the dataset size we are training on. We describe that next for each different dataset used and also corresponding to the \emph{best} models reported in the main paper. We report the total number of epochs and the steps at which the learning rate is decayed.
\myline
\par \noindent Model training iterations for Scaling Data size analysis (Section 4.1 of main paper):
\begin{itemize}
    \item \ImNet permutation 2k: Train for 90 epochs with LR schedule: $150150/150150/100100/50050$.
    \item \ImNetFull permutation 2k: Train for 90 epochs with LR schedule: $1663874/1663874/1109249/554673$.
    \item \YFCCone permutation 2k: Train for 90 epochs with LR schedule: $150150/150150/100100/50050$. 
    \item \YFCCten permutation 2k: Train for 90 epochs with LR schedule: $1171875/1171875/781250/390625$.
    \item \YFCCfifty permutation 2k: Train for 10 epochs only with LR schedule: $651042/651042/434028/217013$.
    \item \YFCCFull permutation 2k: Train for 10 epochs only with LR schedule: $1302083/1302083/868055/434027$.
\end{itemize}

\myline
\par \noindent \textbf The training schedule for the best models (Section 6 of main paper) is as follows:
\begin{itemize}
    \item \ImNet permutation 5k: Train for 90 epochs with LR schedule: $150150/150150/100100/50050$. 
    \item \ImNetFull permutation 5k: Train for 90 epochs with LR schedule: $1663874/1663874/1109249/554673$.
    \item \YFCCFull permutation 10k: Train for 10 epochs with LR schedule: $1302083/1302083/868055/434027$.
\end{itemize}

\subsection{\alexnet\ \colorizationH}
We closely follow the implementation from~\cite{zhang2016colorful} and use the 313 bins and priors provided for training the models. Specifically, we read the original image from the data source, convert the image to Lab, scale the shorter side to 256, randomly crop out $180\times180$ image and randomly flip the image horizontally. Further, following~\cite{zhang2016colorful}, we apply no bias decay for the bias parameter of the model and we also do not apply weight decay to the \texttt{scale} and \texttt{bias} parameter of \spatialbn layers. We train the model on 8-gpus, use a minibatch size of 640, initial learning rate (LR) of 24e-5 with the learning rate dropped by 0.34 at certain interval. We use weight decay 1e-3 and \spatialbn weight decay 0. We use Adam for optimization and beta1 0.9, beta2 0.999 and epsilon 1e-8. The number of training iterations depend on the dataset size we are training on. We describe that next for each different dataset used and also corresponding to the \emph{best} models reported in the main paper.
\myline
\par \noindent Model training iterations used for \colorization models in the main paper:
\begin{itemize}
    \item \ImNet: Train for 28 epochs with LR schedule: $30027/8008/12011/6205$.
    \item \ImNetFull: Train for 112 epochs with LR schedule: $1341749, 347861, 521791, 273319$.
    \item \YFCCone: Train for 28 epochs with LR schedule: $30027/8008/12011/6205$. 
    \item \YFCCten: Train for 56 epochs with LR schedule: $468750/125000/187500/93750$.
    \item \YFCCfifty: Train for 10 epochs only with LR schedule: $1046317/279018/418527/209263$.
    \item \YFCCFull: Train for 15 epochs only with LR schedule: $1265625/328125/492188/257812$.
\end{itemize}

\subsection{\resnetfifty\ \colorizationH}
We closely follow the same setup as for \alexnet described above. We use the 313 bins and priors provided for training the models from~\cite{zhang2016colorful}. We read the original image from the data source, convert the image to Lab, scale the shorter side to 256, randomly crop out $180x180$ image and randomly flip the image horizontally. Further, we apply no bias decay for the bias parameter of the model and we also do not apply weight decay to the \texttt{scale} and \texttt{bias} parameter of \spatialbn layers. We train the model on 8-gpus, use a minibatch size of 640, initial learning rate (LR) of 24e-5 with the learning rate dropped by 0.34 at certain interval. We use weight decay 1e-3 and \spatialbn weight decay 0. We use Adam for optimization and beta1 0.9, beta2 0.999 and epsilon 1e-8. The number of training iterations depends on the dataset size we are training on. We describe that next for each different dataset used and also corresponding to the \emph{best} models reported in the main paper.

\myline
\par \noindent Model training iterations used for \resnetfifty\ \colorization models in the main paper:
\begin{itemize}
    \item \ImNet: Train for 28 epochs with LR schedule: $30027/8008/12011/6205$.
    \item \ImNetFull: Train for 84 epochs with LR schedule: $1006312/260896/391343/204989$.
    \item \YFCCone: Train for 28 epochs with LR schedule: $30027/8008/12011/6205$ 
    \item \YFCCten: Train for 56 epochs with LR schedule: $468750/125000/187500/93750$.
    \item \YFCCfifty: Train for 30 epochs only with LR schedule: $3138951/837054/1255581/627789$.
    \item \YFCCFull: Train for 15 epochs only with LR schedule: $1265625/328125/492188/257812$.
\end{itemize}

\section{Hyperparameters used in Benchmark Tasks}
\label{sec:hyperparam_benchmark}
In this section, we describe hyperparameter settings for various benchmark tasks described in the main paper.

\subsection{Image Classification}
\label{sec:img_cls_appendix}
\myline
\par \noindent \textbf{\VOCseven and \COCO}: We train Linear SVMs on frozen feature representations using LIBLINEAR package~\cite{lin2008liblinear}. We train a linear SVM per class (20 for \VOCseven and 80 for \COCO) for the cost values $C \in 2^{[-19, -4]} \cup  10^{[-7, 2]}$ (\ie 26 $C$ values). We use 3-fold cross-validation to choose the cost parameter per class and then further calculate the mean average precision. To train SVM, we first normalize the features of shape (N, 9k) (where N is number of samples in data and 9k is the resized feature dimension) to have norm=1 along feature dimension. We apply the same normalization step on evaluation data as well. We use the following hyperparameter setting for training using \texttt{LinearSVC} sklearn class. We use \texttt{class\_weight} ratio as 2:1 for positive/negative classes, \texttt{penalty}=l2, \texttt{loss}=squared\_hinge, \texttt{tol}=1e-4, \texttt{dual}=True and \texttt{max\_iter}=2000.

\myline
\par \noindent \textbf{\Places}: We train linear classifiers on frozen feature representations using Nesterov SGD (in~\ref{sec:sgd_dcd}, we discuss the reason for this choice). We freeze the feature representations of various self-supervised networks, resize the features to have dimension 9k and then train linear classifiers. We describe the hyperparameters used for \alexnet and \resnetfifty on both \jigsaw and \colorization approaches.
\begin{enumerate}
    \item \alexnet\ \colorization: We strictly follow~\cite{zhang2016colorful}. Specifically, we train on 8-gpus, use minibatch size of 256, initial learning rate (LR) of 0.01 with the learning rate dropped by factor of 10 at certain interval. We use momentum of 0.9, weight decay 5e-4 and \spatialbn weight decay 0. We do not apply bias decay for the bias parameter of the model and we also do not apply weight decay to the \texttt{scale} and \texttt{bias} parameter of \spatialbn layers. We train for 140k iterations total and use the learning rate schedule of $40k/40k/40k/20$. We read the input image, convert it to Lab, resize the shorter side to 256, randomly crop 227x227 image and apply horizontal flip with 50\% probability. 
    \item \alexnet\ \jigsaw: We follow the settings above and train on 8-gpus, use minibatch size of 256, initial learning rate (LR) of 0.01 with the learning rate dropped by factor of 10 at certain interval. We use momentum of 0.9, weight decay 5e-4 and \spatialbn weight decay 0. We  apply bias decay for the bias parameter of the model and we do not apply weight decay to the \texttt{scale} and \texttt{bias} parameter of \spatialbn layers. We train for 140k iterations total and use the learning rate schedule of $40k/40k/40k/20$. We read the input image, convert it to Lab space, resize the shorter side to 256, randomly crop 227x227 image and apply horizontal flip with 50\% probability.
    \item \resnetfifty\ \colorization:  We closely follow the hyperparameter setting above and train on 8-gpus, use minibatch size of 256, initial learning rate (LR) of 0.01 with the learning rate dropped by factor of 10 at certain interval. We use momentum of 0.9, weight decay 1e-4 and \spatialbn weight decay 0. We do not apply bias decay for the bias parameter of the model. We train for 140k iterations total and use the learning rate schedule of $40k/40k/40k/20k$. We read the input image, convert it to Lab, resize the shorter side to 256, randomly crop 224x224 image and apply horizontal flip with 50\% probability.
    \item \resnetfifty\ \jigsaw:  We closely follow the hyperparameter setting above and train on 8-gpus, use minibatch size of 256, initial learning rate (LR) of 0.01 with the learning rate dropped by factor of 10 at certain interval. We use momentum of 0.9, weight decay 1e-4 and \spatialbn weight decay 0. We apply bias decay for the bias parameter of the model and we do not apply weight decay to the \texttt{scale} and \texttt{bias} parameter of \spatialbn layers. We train for 140k iterations total and use the learning rate schedule of $40k/40k/40k/20k$. We read the input image, convert it to Lab, resize the shorter side to 256, randomly crop 224x224 image and apply horizontal flip with 50\% probability.
\end{enumerate}

\subsection{Low-shot Image Classification}
We train Linear SVMs on \VOCseven and \Places dataset using the exact same setup as in~\ref{sec:img_cls_appendix}. The data sampling technique for various low-shot settings are described in the main paper. 

\subsection{Object Detection}
We follow the same settings as~\cite{girshick2018detectron}. We train on 2-gpus with initial learning rate of 2e-3. For Fast R-CNN, we fine-tune for $22k/8k$ on \VOCseven and for $66k/14k$ on \VOCseventwelve. For Faster R-CNN, we fine-tune for $38k/12k$ on \VOCseven and for $65k/35k$ on \VOCseventwelve. All other hyperparameters are defaults set in \detectron. Note that \detectron default settings use single scale inference with scale value 600.

\subsection{Surface Normal Estimation}
We use the \NYU dataset with the surface normals computed by~\cite{ladicky2014discriminatively}. We use the evaluation metrics from~\cite{fouhey2013data} and the problem formulation from~\cite{wang2015designing}.
\par \noindent \textbf{Problem Setup:} Following~\cite{wang2015designing}, we reduce the problem of surface normal estimation to a classification task. We construct a codebook of size 40 by clustering the surface normals in the \texttt{train} split of \NYU. We then quantize all the surface normals using the codebook and pick the index of the nearest cluster center. This reduces the problem to a 40-way classification problem which we optimize using a multinomial cross-entropy loss. At test time, we predict the distribution over the 40 classes at each pixel location. We convert these per-class distributions into a continuous surface normal prediction by a weighted averaging of the codebook centers with the per-class distribution predictions.

\par \noindent \textbf{Architecture:} We use the PSPNet~\cite{zhao2017pyramid} implementation from~\cite{mitsegmentation}. Specifically we use the ResNet50-dilated backbone encoder and the C1 decoder from their implementation. We only train from \texttt{res5} onwards. We use a learning rate of 2e-2 with a polynomial decay schedule using a power of $0.9$ and a batchsize of 16 across 8 GPUs with Synchronized BatchNorm~\cite{peng2018megdet}. We train all models for 150 epochs and report the best test set performance. The scratch model is trained for 400 epochs and all parameters are updated only for this model. We resize the image with a minimum side of $[300, 375, 450, 525, 600]$ pixels for data augmentation during training (and no left-right flipping).

\subsection{Visual Navigation}
\par \noindent \textbf{Image Features:} We use the implementation and the optimization parameters from~\cite{sax2018mid}. We modify their implementation to use the self-supervised and the supervised \resnetfifty ConvNets to extract features from the images. As their implementation uses an 8 channel feature, we use a random projection of the features from a \resnetfifty. For example, we use a random projection to take the 2048 channel \texttt{res5} features to a 8 channel features. We do \emph{not} train the ConvNet or the random projection matrix. 
\par \noindent \textbf{Agent network architecture:} The Agent uses a recurrent network (GRU~\cite{cho2014learning}) with a state size of 512 dimensions.

\par \noindent \textbf{Optimization:} We use the ADAM~\cite{kingma2014adam} optimizer with a fixed learning rate of $1e\!-\!4$, clipping the $l_2$ norm of the gradient at $0.5$, a rollout size of 512. We use the PPO algorithm~\cite{schulman2017proximal} with a replay buffer size of 10000, value loss weight $1e\!-\!3$, entropy co-efficient $1e\!-\!4$, and a clipping value of $0.1$ for the trust region.

\subsection{Note on Using SGD based Linear Classifiers vs. DCD}
\label{sec:sgd_dcd}
Although, using SGD based Linear Classifiers is a common practice~\cite{zhang2017split} to evaluate representations, we found that optimization hyperparameters can lead to signficantly different results. The SGD based classifiers solve a convex optimization problem, but as also noted in~\cite{kolesnikov2019revisiting}, they can demonstrate a \emph{very} slow convergence. Thus, fine-tuning for larger number of iterations or using a different learning rate decay method (at fixed number of fine-tuning iterations) \etc can give significantly different results.
We obtained more robust results using Dual Coordinate Descent (DCD) as implemented in the LIBLINEAR package~\cite{lin2008liblinear}. Although, this changes the classifier from a logistic regressor to a linear SVM, we believe this setting provides an easy, robust, and fair comparison of image representations and use this setting for the smaller \VOCseven and \COCO datasets.

\section{Hyperparameters used in Legacy Tasks}
\label{sec:hyperparam_legacy}
In this section, we describe hyperparameter settings for various legacy tasks described in Section 7 main paper.

\subsection{\ImNetDataset classification using Linear Classifiers}
We use the exact same setting as described in Section~\ref{sec:img_cls_appendix} for \Places dataset.

\subsection{\VOCseven full fine-tuning}
\label{sec:voc_cls_full}
We use the self-supervised weights to initialize the network and fine-tune the full network on \VOCseven classification task. We use Nesterov SGD for optimization. We describe hyperparameters used in fine-tuning next.
\begin{enumerate}
    \item \alexnet\ \jigsaw: We strictly follow~\cite{zhang2016colorful}. Specifically, we train for 80k iterations on 1-gpu using minibatch size of 16, initial learning rate (LR) of 0.001 with the learning rate dropped by factor of 10 after 10K iterations. We use momentum of 0.9, weight decay 1e-6 and \spatialbn weight decay 1e-4. We do not apply bias decay for the bias parameter of the model and we also do not apply weight decay to the \texttt{scale} and \texttt{bias} parameter of \spatialbn layers. We read the input image, randomly crop 227x227 image and apply horizontal flip with 50\% probability.
    \item \alexnet\ \colorization:  We follow settings above and train for 80k iterations on 1-gpu using minibatch size of 16, initial learning rate (LR) of 0.005 with the learning rate dropped by factor of 10 after 10K iterations. We use momentum of 0.9, weight decay 1e-6 and \spatialbn weight decay 0. We do not apply bias decay for the bias parameter of the model. We read the input image, convert it to Lab, randomly crop 227x227 image and apply horizontal flip with 50\% probability.
    \item \resnetfifty\ \jigsaw: We train for 3000 iterations on 4-gpus using minibatch size of 128, initial learning rate (LR) of 0.01 with the learning rate dropped by factor of 10 after 1600 iterations. We use momentum of 0.9, weight decay 1e-6 and \spatialbn weight decay 0. We do not apply bias decay for the bias parameter of the model  and we also do not apply weight decay to the \texttt{scale} and \texttt{bias} parameter of \spatialbn layers. We read the input image, randomly crop 224x224 image and apply horizontal flip with 50\% probability.
    \item \resnetfifty\ \colorization: We train for 3000 iterations on 4-gpus using minibatch size of 128, initial learning rate (LR) of 0.15 with the learning rate dropped by factor of 10 after 1600 iterations. We use momentum of 0.9, weight decay 1e-6 and \spatialbn weight decay 0. We do not apply bias decay for the bias parameter of the model. We read the input image, convert it to Lab, randomly crop 224x224 image and apply horizontal flip with 50\% probability.
\end{enumerate}

\section{Alternative ways of scaling problem complexity}
\label{sec:extra_problem_complexity}
In the main paper (Section 4.3), we showed how to increase the problem complexity for the \jigsaw task by increasing the size of the permutation set $|\permset|$, and for the \colorization task by changing the number of nearest neighbors $K$ for the soft-encoding. We explore additional ways to increase the problem complexity for the \jigsaw and \colorization methods.

\subsection{\jigsawH: Number of patches $N$}
We increase the number of patches $N$ from 9 (default in~\cite{noroozi2016unsupervised}) to 16. We use $|\permset|=2000$. The input image is resized to $300\times300$ which is divided into a $4\times4$ tiles of size $75\times75$. A $64\times64$ patch is then extracted from each tile randomly to get 16 patches total. We use the same investigation setup as in Section 4 of the main paper - train linear SVMs on the fixed representations for the \VOCseven image classification task. Our results, shown in Table~\ref{tab:jigsaw_patches}, indicate that increasing the number of patches does not result in a higher quality representation. Thus, we only performed further experiments in increasing problem complexity by varying the size of the permutation set $|\permset|$.

\subsection{\colorizationH: Number of color bins $|\colorbins|$}
We increase the size of the color bins $|\colorbins|$ that are used to quantize the color space (see Section 3.2 of the main paper). This increases the number of colors the ConvNet predicts for the \colorization problem. We evaluate the quality of the features by transfer learning on the \Places dataset (same setup as in Section 5 of the main paper). As Table~\ref{tab:color_bins} shows, using $|\colorbins| \in [313, 262]$ gives the best results. Thus, we use 313 bins in the main paper which is also the default in~\cite{zhang2016colorful}. 

We also experimented with the bandwidth of the Gaussian used to compute the soft-encoding $\imageabBinned^\numsoftbins$ but did not see any significant improvements.

\section{Additional Results}
\label{sec:extra_results}

\subsection{Image Classification using Linear Classifiers on \ImNetDataset}
Similar to Section 7 of the main paper, we train linear classifiers on frozen representations from different layers of a \resnetfifty model in Table~\ref{tab:rn50_linear_in1k}. We strictly follow the setup from Zhang \etal~\cite{zhang2017split} and compare against earlier works that also use \resnetfifty for self-supervised pre-training.
\begin{table*}[!]
\centering
\setlength{\tabcolsep}{0.65em}\scalebox{0.8}{
\begin{tabular}{l|ccccc}
& \multicolumn{5}{c}{\textbf{\ImNet}}\\
 \textbf{Method} & \textbf{\mytexttt{layer1}} & \textbf{\mytexttt{layer2}} & \textbf{\mytexttt{layer3}} & \textbf{\mytexttt{layer4}} & \textbf{\mytexttt{layer5}}\\
\shline
\shline
\resnetfifty \ImNet Supervised & 11.6 & 33.3 & 48.7 & 67.9 & 75.5\\
\resnetfifty \Places Supervised & 13.2 & 31.7 & 46.0 & 58.2 & 51.7\\
\thinline
\resnetfifty Random & 9.6 & 13.7 & 12.0 & 8.0 & 5.6\\
\thinline
\resnetfifty (Kolesnikov \etal)~\cite{kolesnikov2019revisiting}$^\dagger$ & -- & -- & -- & 47.7 & -- \\
\resnetfifty (NPID)~\cite{wu2018unsupervised}$^\triangleleft$ & 15.3 & 18.8 & 24.9 & 40.6 & 54.0 \\
\thinline
\resnetfifty\ \jigsaw \ImNet & \textbf{12.4} & 28.0 & \underline{39.9} & 45.7 & 34.2 \\
\resnetfifty\ \jigsaw \ImNetFull & 7.9 & \textbf{30.2} & 39.0 & 46.3 & 35.9 \\
\resnetfifty\ \jigsaw \YFCCFull & 7.9 & 28.2 & \textbf{41.3} & 48.3 & 34.7 \\
\resnetfifty\ \mytexttt{Coloriz.} \ImNet & 10.2 & 24.1 & 31.4 & 39.6 & 35.2 \\
\resnetfifty\ \mytexttt{Coloriz.} \ImNetFull & 10.1 & 27.0 & 37.8 & \textbf{49.4} & \textbf{46.2} \\
\resnetfifty\ \mytexttt{Coloriz.} \YFCCFull & 10.4 & 25.9 & 37.7 & 47.8 & 41.1 \\
\shline
\shline
\end{tabular}}
\vspace{-0.1in}
\caption{\textbf{ResNet-50 top-1 center-crop accuracy for linear classification on the \ImNet dataset.}. Numbers with $^\dagger$ are with $10-20\times$ longer fine-tuning and are reported on unofficial \ImNet validation split. Numbers with $^\triangleleft$ use different fine-tuning procedure. All other models follow the setup from Zhang \etal~\cite{zhang2017split}.}
\label{tab:rn50_linear_in1k}
\end{table*}

\subsection{Image Classification using Linear SVM on \VOCseven}
Complete results of \textbf{Section 6.1} in the main paper. We report results using the \alexnet model on the \VOCseven dataset in Table~\ref{tab:linearsvm_alexnet_voc}.
\begin{table}[!]
\centering
\setlength{\tabcolsep}{0.2em}\scalebox{0.75}{
\begin{tabular}{l|ccccc}
\textbf{Method} & \textbf{\mytexttt{layer1}} & \textbf{\mytexttt{layer2}} & \textbf{\mytexttt{layer3}} & \textbf{\mytexttt{layer4}} & \textbf{\mytexttt{layer5}}\\
\shline
\alexnet \ImNet Supervised & 34.9 & 51.8 & 59.5 & 64.6 & 68.0 \\
\alexnet \Places Supervised & 34.4 & 51.7 & 59.4 & 63.2 & 65.7 \\
\alexnet Random & 8.5 & 7.9 & 8.0 & 7.8 & 7.9 \\
\thinline
\alexnet\ \jigsaw \ImNet & 35.6 & 48.3 & 53.2 & 53.5 & 49.5 \\
\alexnet\ \jigsaw \ImNetFull & 36.1 & 49.0 & 53.9 & 54.3 & 48.3 \\
\alexnet\ \jigsaw \YFCCFull & 35.9 & 49.1 & 54.7 & 55.4 & 49.7 \\
\thinline
\alexnet\ \colorization \ImNet & 31.7 & 43.4 & 47.6 & 50.3 & 51.6 \\
\alexnet\ \colorization \ImNetFull & 31.5 & 45.6 & 50.8 & 54.6 & 55.7 \\
\alexnet\ \colorization \jigsaw \YFCCFull & 32.7 & 46.0 & 51.7 & 54.3 & 55.1\\
\shline
\end{tabular}}
\caption{AlexNet \textbf{linear SVM} classification on the \VOCseven dataset.}
\label{tab:linearsvm_alexnet_voc}
\end{table}

\subsection{Image Classification using Linear SVM on \COCO}
Complete results of \textbf{Section 6.1} in the main paper. We report results on the \COCO dataset in Table~\ref{tab:linearsvm_coco}.
\begin{table}[!]
\centering
\setlength{\tabcolsep}{0.2em}\scalebox{0.75}{
\begin{tabular}{l|ccccc}
\textbf{Method} & \textbf{\mytexttt{layer1}} & \textbf{\mytexttt{layer2}} & \textbf{\mytexttt{layer3}} & \textbf{\mytexttt{layer4}} & \textbf{\mytexttt{layer5}}\\
\shline
\resnetfifty\ \jigsaw \ImNet & 19.6 & 33.9 & 41.9 & 47.3 & 41.1\\
\resnetfifty\ \jigsaw \ImNetFull & 14.7 & 34.9 & 43.4 & 52.1 & 45.5\\
\resnetfifty\ \jigsaw \YFCCFull & 14.7 & 34.2 & 43.7 & 52.2 & 44.4\\
\thinline
\resnetfifty \ImNet Supervised & 17.5 & 35.5 & 45.8 & 60.5 & 68.5\\
\shline
\alexnet\ \jigsaw \ImNet & 25.8 & 35.0 & 38.7 & 38.6 & 34.6\\
\alexnet\ \jigsaw \ImNetFull & 25.9 & 35.6 & 39.7 & 39.3 & 34.1\\
\alexnet\ \jigsaw \YFCCFull & 25.8 & 35.8 & 40.2 & 40.1 & 33.9\\
\thinline
\alexnet \ImNet Supervised & 24.4 & 37.9 & 43.4 & 46.5 & 47.6\\
\shline
\end{tabular}}
\caption{\textbf{Linear SVM} classification on the \COCO dataset.}
\label{tab:linearsvm_coco}
\end{table}

\subsection{Image Classification using Full fine-tuning on \VOCseven}
We perform full fine-tuning (hyperparameters in~\cref{sec:voc_cls_full}) of the self-supervised networks for the \VOCseven classification task. We provide results for the \resnetfifty model in Table~\ref{tab:fullft_rn50} and for \alexnet in Table~\ref{tab:fullft_voc2007_alexnet}. The \resnetfifty model matches the performance of a supervised pre-trained \Places model on the \VOCseven classification task. We note that obtaining a comparable evaluation setting to prior work using \alexnet is difficult because of differences in fine-tuning schedule and weight re-scaling methods.

\begin{table}[!t]
\centering
\setlength{\tabcolsep}{0.6em}\scalebox{0.85}{
\begin{tabular}{l|c}
\textbf{Method} & \textbf{\VOCseven}\\
\shline
\shline
\resnetfifty \ImNet Supervised & 90.3\\
\resnetfifty \Places Supervised & 86.9\\
\thinline
\resnetfifty Random$^*$ & 48.4\\
\thinline
\resnetfifty\ \jigsaw \ImNet & 73.9\\
\resnetfifty\ \jigsaw \ImNetFull & \textbf{83.8}\\
\resnetfifty\ \jigsaw \YFCCFull & 82.7\\
\resnetfifty\ \mytexttt{Coloriz.} \ImNet & 67.9\\
\resnetfifty\ \mytexttt{Coloriz.} \ImNetFull & 75.0\\
\resnetfifty\ \mytexttt{Coloriz.} \YFCCFull & 75.3\\
\shline
\shline
\end{tabular}}
\caption{\textbf{\resnetfifty Full fine-tuning image classification (mAP scores) for \VOCseven}. All models are trained using the same setup and we report center crop numbers. Random initialization baseline (denoted with $^*$) is trained for $4\times$ longer fine-tuning schedule on \VOCseven. }
\label{tab:fullft_rn50}
\end{table}
\begin{table}[!t]
\centering
\setlength{\tabcolsep}{0.7em}\scalebox{0.85}{
\begin{tabular}{l|c}
\textbf{Method} & \textbf{\VOCseven}\\
\shline
\alexnet \ImNet Supervised & 79.9\\
\alexnet \Places Supervised & 75.7\\
\thinline
\alexnet Random$^\ddagger$ & 53.3\\
\thinline
\alexnet (Context)~\cite{doersch2015unsupervised}$^\triangleleft$ & 65.3\\
\alexnet (SplitBrain)~\cite{zhang2017split}$^\triangleleft$ & 67.1\\
\alexnet (Counting)~\cite{noroozi2017representation}$^\triangleleft$ & 67.7\\
\alexnet (Rotation)~\cite{gidaris2018unsupervised}$^{\triangleleft,*}$ &  72.9 \\
\alexnet (DeepCluster)~\cite{caron2018deep}$^\dagger$ & 70.4\\
\thinline
\alexnet\ \jigsaw \ImNet & 58.5\\
\alexnet\ \jigsaw \ImNetFull & 63.7\\
\alexnet\ \jigsaw \YFCCFull & 63.0\\
\alexnet\ \mytexttt{Coloriz.} \ImNet & 61.9\\
\alexnet\ \mytexttt{Coloriz.} \ImNetFull & \textbf{66.7}\\
\alexnet\ \mytexttt{Coloriz.} \YFCCFull & 65.5\\
\shline
\end{tabular}}
\caption{\textbf{\alexnet Full fine-tuning image classification (mAP scores) for \VOCseven}: We report 10-crop numbers as in~\cite{zhang2017split}. Method with $^\dagger$ uses a different fine-tuning schedule, $^\triangleleft$ uses weight re-scaling, $^*$ we could not determine exact fine-tuning details. Numbers with $^\ddagger$ taken from~\cite{zhang2017split}. We note that drawing consistent comparisons with (and among) prior work is difficult because of differences in the fine-tuning procedure and thus present these results only for the sake of completeness.}
\label{tab:fullft_voc2007_alexnet}
\end{table}

\subsection{Layerwise Results for Low-shot}
We report results on low-shot classification (Section 6.2 in the main paper) in Figure~\ref{fig:lowshot_layerwise}. We show the results for both the \colorization and \jigsaw self-supervised methods for a \resnetfifty model.

\begin{table*}[!t]
    \centering
    \begin{tabular}{cc}
        \includegraphics[width=0.45\textwidth]{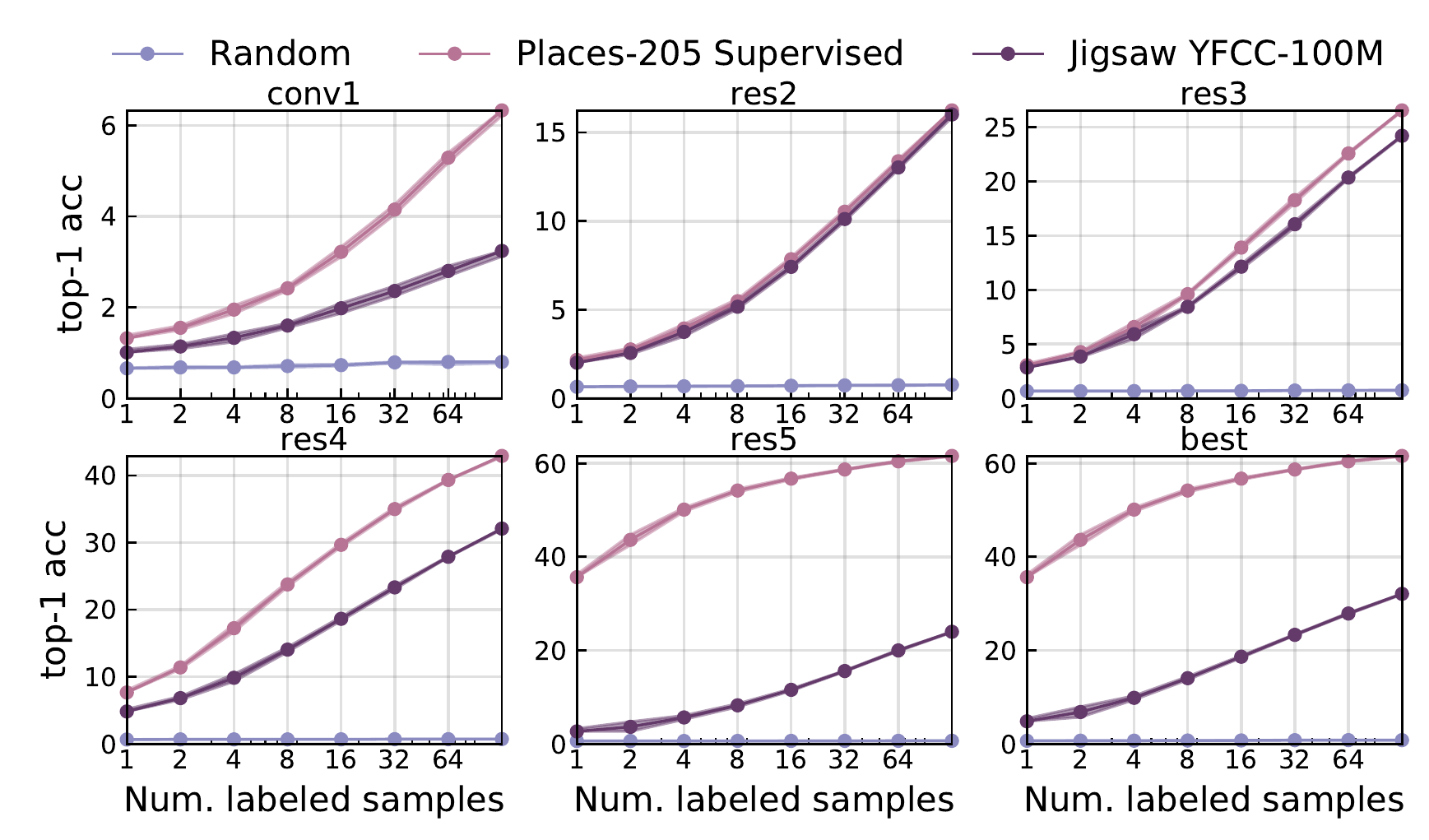} & \includegraphics[width=0.45\textwidth]{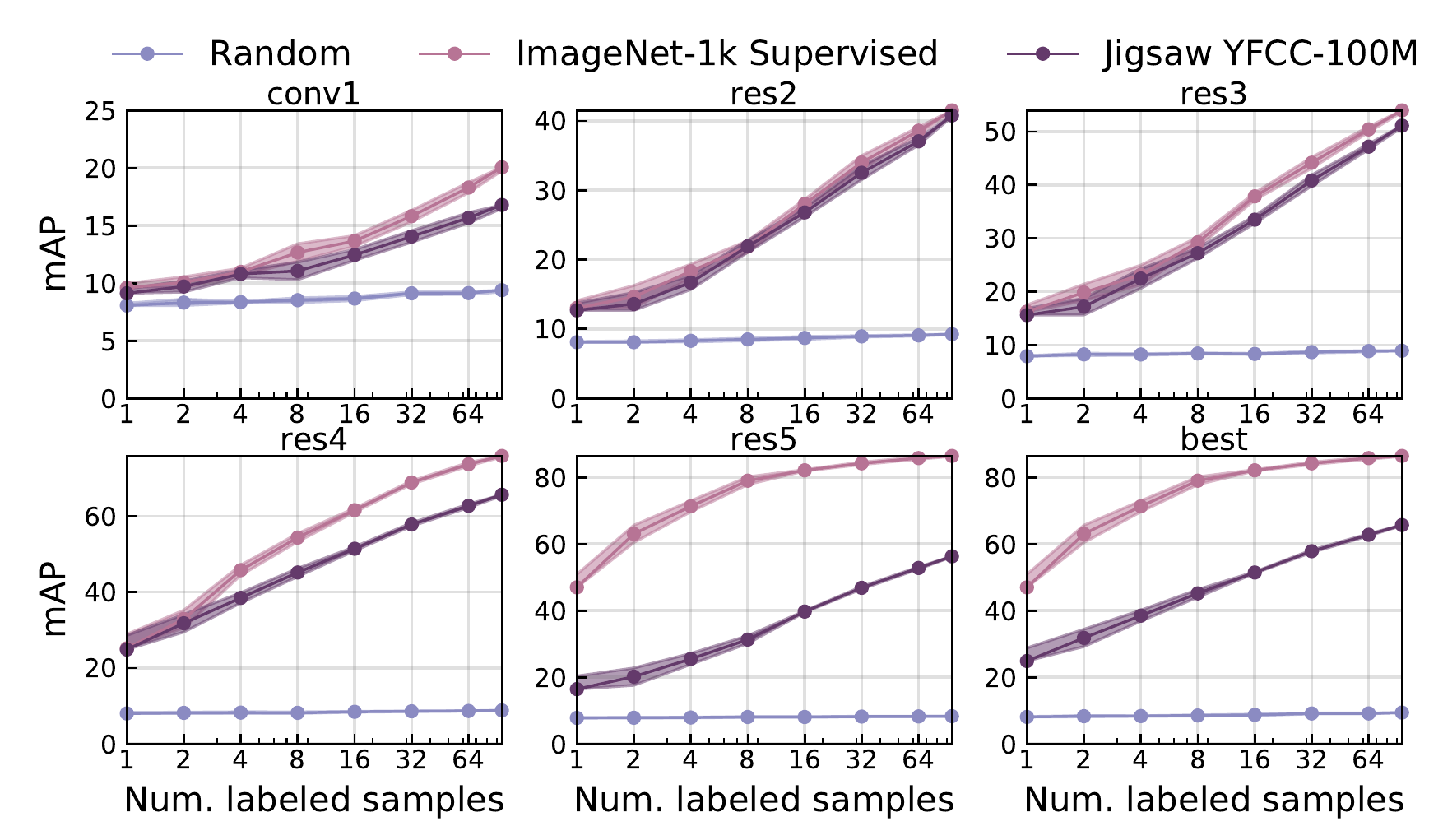} \\
        (a) \jigsaw \Places & (b) \jigsaw \VOCseven \\
        \includegraphics[width=0.45\textwidth]{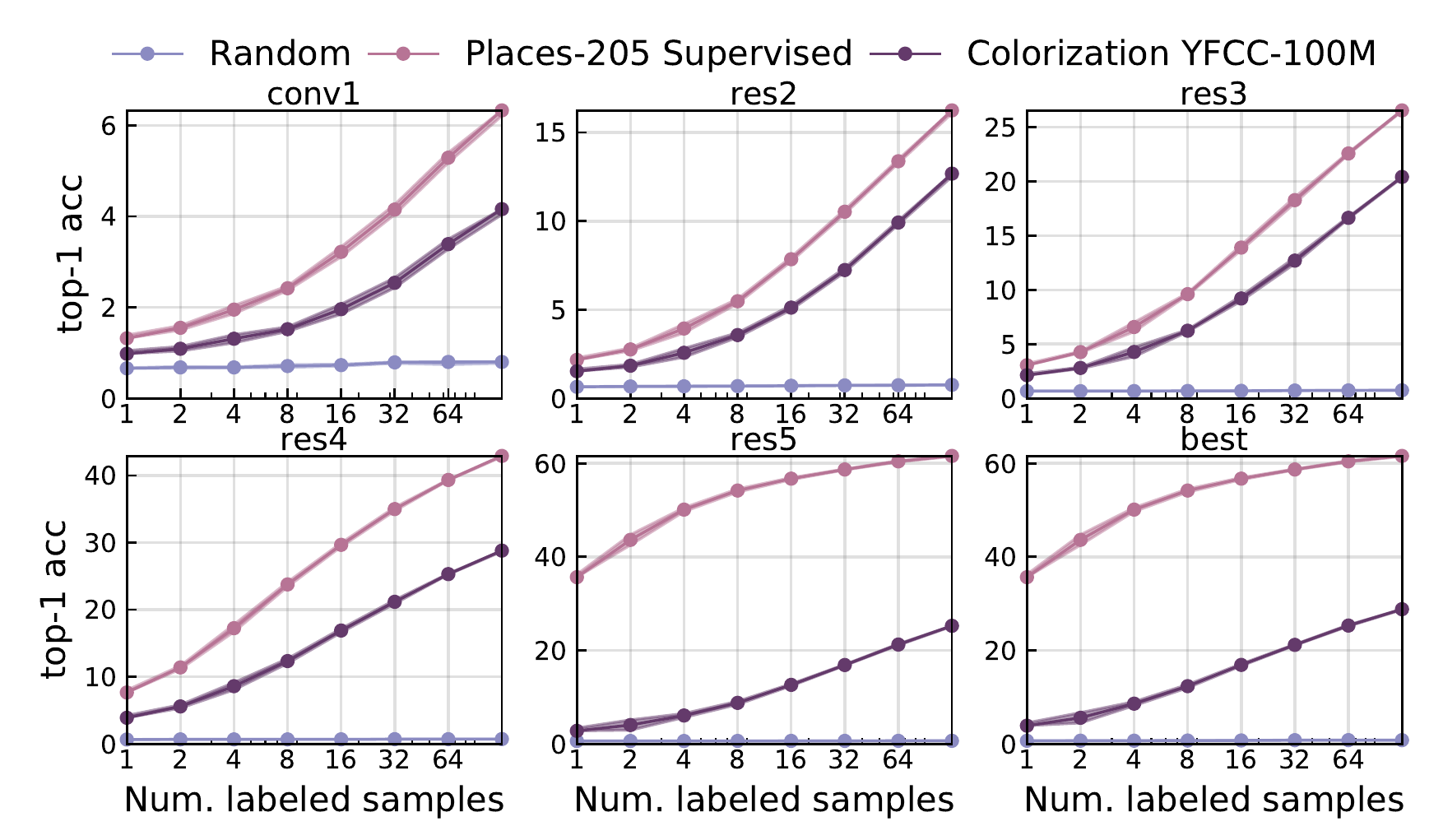} & \includegraphics[width=0.45\textwidth]{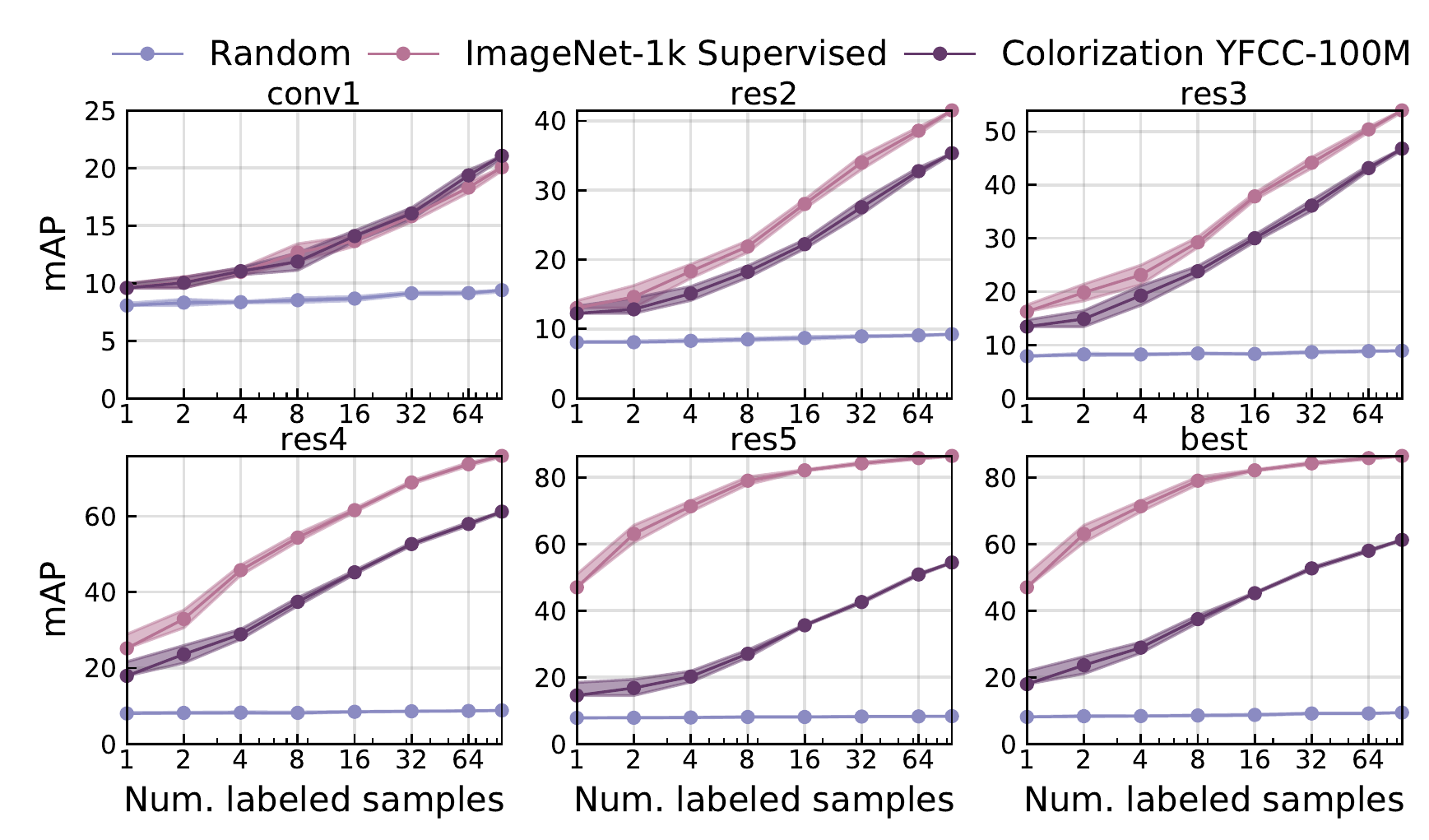} \\
        (a) \colorization \Places & (b) \colorization \VOCseven \\
    \end{tabular}
    \captionof{figure}{Low-shot Classification (layerwise) results on the using a \resnetfifty. These are additional results following the same setup as in Section 6.2 of the main paper. We train linear classifiers (SVMs) on layer-wise representations.}
    \label{fig:lowshot_layerwise}
\end{table*}

\subsection{Surface Normal Estimation using a \resnetfifty\ Colorization model}
We report results  on the Surface Normal Estimation task using a \resnetfifty self-supervised model on the \colorization method (setup from Section 6.4 of the paper). These results are in Table~\ref{tab:surface_normal_frozen_colorization}.
\begin{table}[!h]
\setlength{\tabcolsep}{0.12em}
\centering
\scalebox{0.75}{
    \begin{tabular}{@{}l|cccccc@{}}
    & \multicolumn{2}{c}{\textbf{Angle Distance}} && \multicolumn{3}{c}{\textbf{Within $t^\circ$}} \\
    \textbf{Initialization} & Mean & Median && 11.25 & 22.5 & 30 \\
    & \multicolumn{2}{c}{\textbf{(Lower is better)}} && \multicolumn{3}{c}{\textbf{(Higher is better)}} \\
    \shline
    \resnetfifty\ \ImNet supervised & 26.4 &	17.1 && 36.1 & 59.2 & 68.5 \\
    \resnetfifty\ \Places supervised & 23.3 & 14.2 && 41.8 & 65.2 & 73.6 \\
     \thinline
    \resnetfifty\ Scratch & 26.3	& 16.1 &&	37.9 &	60.6 &	69.0 \\
    \thinline
    \colorization \YFCCFull & 28.5 & 22.6 && 28.1 & 49.9 & 61.4 \\
    \colorization \ImNetFull & 27.1 & 19.5 && 32.4 & 55.1 & 65.6 \\
    \colorization \ImNet & 29.3 & 21.0 && 30.2 & 52.4 & 62.8 \\
    \shline
    \end{tabular}}
\vspace{-0.08in}
\caption{\textbf{Surface Normal Estimation on the \NYU dataset}. We train \resnetfifty from \texttt{res5} onwards and freeze the \texttt{conv} body below.}
\label{tab:surface_normal_frozen_colorization}

\end{table}

\subsection{Faster R-CNN results on \VOCseven and \VOCseventwelve}
Similar to Section 6.3 of the main paper, we freeze the \texttt{conv} body for all the models. We train the ROI-heads and the classifier (\texttt{res5} onwards). We report these results in Table~\ref{tab:detection-faster-Fbody}. For all the methods (including supervised and self-supervised), we use a slightly longer fine-tuning schedule of $38k/12k$ for \VOCseven and $65k/35k$ for \VOCseventwelve. All other parameters are kept the same as in \detectron~\cite{girshick2018detectron}.

\begin{table}[!]
\centering
\setlength{\tabcolsep}{0.7em}\scalebox{0.80}{
\begin{tabular}{l|cc}
\textbf{Method} & \textbf{\VOCseven} & \textbf{\VOCseventwelve}\\
\shline
\resnetfifty \ImNet Supervised & 67.1 & 68.3\\
\thinline
\resnetfifty
\resnetfifty\ \jigsaw \ImNetFull & 62.7 & 64.8\\
\resnetfifty\ \jigsaw \YFCCFull & 56.9 & 59.8 \\
\shline
\end{tabular}}
\vspace{-0.08in}
\caption{\textbf{Detection mAP for frozen \texttt{conv} body} on \VOCseven and \VOCseventwelve using Faster R-CNN with ResNet-50-C4. We freeze the \texttt{conv} body for all models.}
\label{tab:detection-faster-Fbody}
\end{table}

\par \noindent \textbf{Full fine-tuning:} We evaluate in the \emph{full} fine-tuning setting to draw comparisons with~\cite{wu2018unsupervised}. We use the default parameters in \detectron~\cite{girshick2018detectron} for this setting. Our \resnetfifty\ \jigsaw\ \ImNetFull model fine-tuned on \VOCseven \texttt{trainval} obtains $68.9$ mAP on \VOCseven \texttt{test} compared to $65.4$ reported in~\cite{wu2018unsupervised}. We show full results on this setting in Table~\ref{tab:detection-faster-full}.

\begin{table}[!]
\centering
\setlength{\tabcolsep}{0.7em}\scalebox{0.80}{
\begin{tabular}{l|cc}
\textbf{Method} & \textbf{\VOCseven} & \textbf{\VOCseventwelve}\\
\shline
\resnetfifty \ImNet Supervised & 70.9 & 76.4 \\
\thinline
\resnetfifty
\resnetfifty\ \jigsaw \ImNet & 64.5  & 67.3 \\
\resnetfifty\ \jigsaw \ImNetFull & 68.9 & 75.3 \\
\resnetfifty\ \jigsaw \YFCCFull & 66.4 & 73.9 \\
\shline
\end{tabular}}
\vspace{-0.08in}
\caption{\textbf{Detection mAP with full fine-tuning} on \VOCseven and \VOCseventwelve using Faster R-CNN with ResNet-50-C4. We freeze the \texttt{conv} body for all models.}
\label{tab:detection-faster-full}
\end{table}

\begin{table}[]
\centering
\setlength{\tabcolsep}{0.2em}\scalebox{0.7}{
\begin{tabular}{c|ccccc}
& \multicolumn{5}{c}{\textbf{\VOCseven}}\\
 \textbf{Number of patches ($N$)} & \textbf{\mytexttt{layer1}} & \textbf{\mytexttt{layer2}} & \textbf{\mytexttt{layer3}} & \textbf{\mytexttt{layer4}} & \textbf{\mytexttt{layer5}}\\
\shline
9 & 26.7 & 44.6 & 53.5 & 64.1 & 55.5 \\
16 & 31.9 & 42.0 & 48.1 & 49.8 & 37.9 \\
\shline
\end{tabular}}
\vspace{-0.1in}
\caption{\textbf{Varying number of patches $N$ for a \resnetfifty on \jigsaw}. We increase the problem complexity of the \jigsaw method by increasing the number of patches from 9 (default in~\cite{noroozi2016unsupervised}) to 16. We keep the size of the permutation set fixed at $|\permset|=2000$. We report the performance of training a linear SVM on the fixed features for the \VOCseven image classification task. We do not see an improvement by increasing the number of patches.}
\label{tab:jigsaw_patches}
\end{table}
\begin{table}[]
\centering
\setlength{\tabcolsep}{0.2em}\scalebox{0.7}{
\begin{tabular}{c|ccccc}
& \multicolumn{5}{c}{\textbf{\Places}}\\
 \textbf{Number of bins ($|\colorbins|$)} & \textbf{\mytexttt{layer1}} & \textbf{\mytexttt{layer2}} & \textbf{\mytexttt{layer3}} & \textbf{\mytexttt{layer4}} & \textbf{\mytexttt{layer5}}\\
\shline

968 & 14.5 & 27.8 & 32.0 & 33.1 & 36.3 \\
313 & 14.5 & 27.5 & 32.8 & 37.6 & 34.6 \\
262 & 14.8 & 27.9 & 32.5 & 38.6 & 36.5 \\
124 & 14.5 & 26.6 & 30.7 & 27.7 & 35.3 \\
76 & 15.6 & 28.1 & 32.7 & 33.6 & 36.5 \\

\shline
\end{tabular}}
\vspace{-0.1in}
\caption{\textbf{Varying number of colorbins $|\colorbins|$ for a \resnetfifty on \colorization}. We increase the problem complexity for the \colorization method by increasing the number of colors ($|\colorbins|$) the ConvNet must predict. We evaluate the feature representation by training linear classifiers on the fixed features. We report the top-1 center crop accuracy on the \Places dataset.}
\label{tab:color_bins}
\end{table}

\end{document}